\documentclass[sigconf]{acmart}

\AtBeginDocument{%
  }
\usepackage{enumitem}

\newcommand{\eg}{{\emph{e.g.,}\ }}
\newcommand{\ie}{{\emph{i.e.,}\ }}
\newcommand{\etc}{{\emph{etc.}\ }}
\newcommand{\wrt}{{\emph{w.r.t,}\ }}

\newcommand{\name}[0]{smileGeo}

\usepackage{array}

\usepackage{booktabs}
\usepackage{csvsimple}
\usepackage{multirow}
\usepackage{makecell}
\makeatother

\usepackage{balance}

\usepackage{tikz}
\usepackage{pgfplots}
\pgfplotsset{width=8cm,compat=1.13}
\usepackage{arydshln} % dashed line
\usetikzlibrary{arrows,shapes,shadows,snakes,positioning,automata,backgrounds,patterns}

\usepackage{graphicx}
\usepackage{subfigure}

\usepackage{caption}
\usepackage{subcaption}

\usepackage{algorithm, algorithmic}

\usepackage{enumitem}

\usepackage{amsmath}
\usepackage{bbm}

\usepackage{caption}
\usepackage{xcolor}
\usepackage{color}

\definecolor{mycolor51}{RGB}{84,89,105}    % 1 black
\definecolor{mycolor52}{RGB}{164,117,125}  % 2 brown
\definecolor{mycolor53}{RGB}{231,152,124} % 3 orange
\definecolor{mycolor54}{RGB}{139,145,182}   % 4 blue
\definecolor{mycolor55}{RGB}{119,113,164} % 5 purple

\definecolor{mycolor55light}{RGB}{171,164,195}

\definecolor{mycolor41}{RGB}{54,80,131}
\definecolor{mycolor42}{RGB}{183,131,175}
\definecolor{mycolor43}{RGB}{245,166,115}
\definecolor{mycolor44}{RGB}{252,219,114}

\definecolor{mycolor42light}{RGB}{206,178,203}
\definecolor{mycolor42light2}{RGB}{222,204,221}
\definecolor{mycolor43light}{RGB}{245,201,169}
\definecolor{mycolor44light}{RGB}{249,230,175}

\definecolor{mycolor61}{RGB}{17,50,93}
\definecolor{mycolor62}{RGB}{54,80,131}
\definecolor{mycolor63}{RGB}{115,107,157}
\definecolor{mycolor64}{RGB}{183,131,175}
\definecolor{mycolor65}{RGB}{245,166,115}
\definecolor{mycolor66}{RGB}{252,219,114}

\definecolor{mycolor31}{RGB}{189,142,192}
\definecolor{mycolor32}{RGB}{111,128,190}
\definecolor{mycolor33}{RGB}{244,126,98}

\definecolor{mycolor01}{RGB}{103,0,13}
\definecolor{mycolor02}{RGB}{165,15,21}
\definecolor{mycolor03}{RGB}{203,24,29}
\definecolor{mycolor04}{RGB}{239,59,44}
\definecolor{mycolor05}{RGB}{251,106,74}
\definecolor{mycolor06}{RGB}{252,146,114}
\definecolor{mycolor07}{RGB}{252,187,161}
\definecolor{mycolor08}{RGB}{254,224,210}

\definecolor{mycolor09}{RGB}{222,235,247}
\definecolor{mycolor10}{RGB}{198,219,239}
\definecolor{mycolor11}{RGB}{158,202,225}
\definecolor{mycolor12}{RGB}{107,174,214}
\definecolor{mycolor13}{RGB}{66,146,198}
\definecolor{mycolor14}{RGB}{33,113,181}
\definecolor{mycolor15}{RGB}{8,81,156}
\definecolor{mycolor16}{RGB}{8,48,107}

\acmYear{2025}
\acmConference[WWW '25]{Proceedings of the Web Conference 2025}{April 28--May 2, 2025}{Sydney, Australia}
\acmBooktitle{Proceedings of the Web Conference 2025 (WWW '25), April 28--May 2, 2025, Sydney, Australia}
\begin{document}

% 调整间距
% \setlength{\intextsep}{9pt} % 调整图形与周围文本之间的间距
% \setlength{\abovecaptionskip}{5pt} % 调整标题上方的间距
% \setlength{\belowcaptionskip}{0pt}  % 调整标题下方的间距

\title{Swarm Intelligence in Geo-Localization: A Multi-Agent Large Vision-Language Model Collaborative Framework}


\author{Xiao Han}
\email{hahahenha@gmail.com}
% \authornotemark[2]
\orcid{0000-0002-3478-964X}
\affiliation{%
  \institution{City University of Hong Kong}
  % \streetaddress{83 Tat Chee Avenue}
  \city{Hong Kong}
  \country{China}
  % \postcode{999077}
}

\author{Chen Zhu}
\email{zc3930155@gmail.com}
\authornotemark[1]
% \orcid{0000-000x-xxxx-xxxx}
\affiliation{%
  \institution{University of Science and Technology of China}
  % \streetaddress{xxxx}
  \city{Hefei}
  % \state{Beijing}
  \country{China}
  % \postcode{xxxxxx}
}

\author{Xiangyu Zhao}
\email{xianzhao@cityu.edu.hk}
\authornotemark[1]
\affiliation{
  \institution{City University of Hong Kong}
  % \streetaddress{83 Tat Chee Avenue}
  \city{Hong Kong}
  \country{China}
  % \postcode{999077}
}

\author{Hengshu Zhu}
\email{hengshuzhu@hkust-gz.edu.cn}
\authornotemark[1]
% \orcid{0000-000x-xxxx-xxxx}
\affiliation{%
  \institution{Hong Kong University of Science and Technology (Guangzhou)}
  % \streetaddress{xxxx}
  \city{Guangzhou}
  % \state{}
  \country{China}
  % \postcode{xxxxxx}
}

\renewcommand{\shortauthors}{Xiao Han et al.}


% a session-based framework, \name, is developed, aiming to accurately reflect users' real-time job preferences through behavioral-semantic fusion learning.

\begin{abstract}

Visual geo-localization demands in-depth knowledge and advanced reasoning skills to associate images with precise real-world geographic locations.
% In general, 
Existing image database retrieval methods are limited by the impracticality of storing sufficient visual records of global landmarks.
% Additionally, in agent-based systems, conflicting conclusions among agents typically necessitate third-party intervention to reach a unified decision.
Recently, Large Vision-Language Models (LVLMs) have demonstrated the capability of geo-localization through Visual Question Answering (VQA), enabling a solution that does not require external geo-tagged image records.
% do we need the following line?
However, the performance of a single LVLM is still limited by its intrinsic knowledge and reasoning capabilities.
% Inspired by their strong capabilities, in this paper, we introduce ...
To address these challenges, we introduce \name, a novel visual geo-localization framework that leverages multiple Internet-enabled LVLM agents operating within an agent-based architecture. By facilitating inter-agent communication, \name\ integrates the inherent knowledge of these agents with additional retrieved information, enhancing the ability to effectively localize images.
Furthermore, our framework incorporates a dynamic learning strategy that optimizes agent communication, reducing redundant interactions and enhancing overall system efficiency.
To validate the effectiveness of the proposed framework, we conducted experiments on three different datasets, and the results show that our approach significantly outperforms current state-of-the-art methods.
The source code is available at https://anonymous.4open.science/r/ViusalGeoLocalization-F8F5.

\noindent \textbf{Relevance Statement:} This paper focuses on invoking search, inferring the geo-locations of images through discussion and analysis of the retrieved information among multiple LVLMs, and responding in the form of natural language. The designed method in this paper can assist search and retrieval-augmented AI applications.

\end{abstract}
\keywords{LLM Agent, Graph Neural Networks, Retrieval-augmented LLM}

\begin{CCSXML}
<ccs2012>
   <concept>
       <concept_id>10002951.10003227.10003251</concept_id>
       <concept_desc>Information systems~Multimedia information systems</concept_desc>
       <concept_significance>500</concept_significance>
       </concept>
   <concept>
       <concept_id>10010147.10010178.10010219.10010221</concept_id>
       <concept_desc>Computing methodologies~Intelligent agents</concept_desc>
       <concept_significance>500</concept_significance>
       </concept>
 </ccs2012>
\end{CCSXML}

\ccsdesc[500]{Information systems~Multimedia information systems}
\ccsdesc[500]{Computing methodologies~Intelligent agents}


% \received{8 February 2024}
% \received[revised]{5 April 2024}
% \received[accepted]{16 May 2024}

\maketitle

\section{Introduction}

Visual geo-localization, referred to the task of estimating geographical identification for a given image,
% also known as geotagging\footnote{\href{https://en.wikipedia.org/wiki/Geotagging}{https://en.wikipedia.org/wiki/Geotagging}}, 
is vital in various fields, such as 
analyzing historical human mobility patterns \cite{DBLP:conf/asunam/HuangC19,DBLP:journals/mta/LuoJYG11,DBLP:journals/pami/ArandjelovicGTP18,DBLP:journals/ijcv/ZaffarGMKFME21,DBLP:journals/pami/ToriiASOP18} and providing location-aware, city-level attraction recommendations \cite{DBLP:conf/icra/ChenJSULSRM17,DBLP:journals/ral/ChenLSGC18,DBLP:conf/iros/ChenMSC17,DBLP:journals/ijrr/GargSM22,DBLP:journals/ral/HauslerJM19,DBLP:journals/trob/KhaliqECMM20}.
In general, accurately geo-localizing images without relying on localization-based metadata (e.g., GPS tags) is a complex task that demands extensive geospatial knowledge and advanced reasoning capabilities.
Traditional methods \cite{DBLP:journals/envsoft/ElQadiLDD20,campbell2006vision,DBLP:journals/tip/DengZGQGC20,DBLP:journals/tie/ZhangDCBPCC18} typically formulate it as an image retrieval problem where to geo-localize the given image by retrieving similar images with known geographical locations.
Thus, their effectiveness is limited by the scope and quality of the geo-tagged image records.

\begin{figure}[t]
\centering
\includegraphics[width=\linewidth]{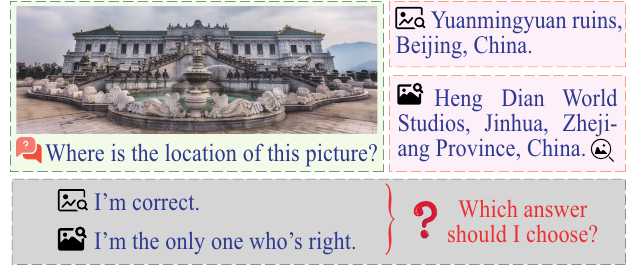}
\caption{A toy example in the traditional agent-based retrieval framework.}
\label{fig:1}
\end{figure}

A straightforward approach to mitigate the limitation of these databases is to deploy an agent-based framework, leveraging swarm intelligence across multiple retrieval systems \cite{DBLP:journals/corr/abs-2303-12712,DBLP:conf/nips/SchaefferMK23}.
However, integrating diverse and independent traditional retrieval systems as agents within a unified framework presents significant challenges without human intervention.
For instance, as illustrated in Figure~\ref{fig:1}, two distinct retrieval systems may provide different answers for the same input image, making it difficult to decide on the correct response without third-party mediation.
This challenge highlights the inherent difficulty of coordinating independent systems within a swarm intelligence framework.

In parallel with agent-based approaches,
recent advancements in Large Vision-Language Models (LVLMs) have opened up new possibilities for multi-modal tasks, such as Visual Question Answering (VQA) \cite{feng2024large,wang2023cogvlm}.
LVLMs offer an innovative solution to visual geo-localization without relying on external geo-tagged image records.
Additionally, enabling LVLMs to autonomously query network search interfaces for information retrieval can further enhance their capabilities \cite{xu2024reverse}.
However, while individual LVLMs possess strong reasoning abilities, they still struggle with fine-grained recognition across diverse and complex scenes \cite{DBLP:journals/corr/abs-2310-01444,DBLP:journals/corr/abs-2312-08914,DBLP:journals/corr/abs-2310-02170}.

To address the limitations of both traditional retrieval systems and individual LVLMs, we propose a novel multi-LVLM agent framework, named \underline{\textbf{s}}war\underline{\textbf{m}}
\underline{\textbf{i}}ntel\underline{\textbf{l}}ig\underline{\textbf{e}}nce
\underline{\textbf{Geo}}-localization (\textbf{\name}).
This framework leverages the swarm intelligence of multiple LVLMs,
each equipped with network retrieval capabilities,
to collaboratively and efficiently geo-localize images.
Specifically, for a given image, the framework initially selects $K$ appropriate LVLM agents to serve as answer agents responsible for conducting the initial location analysis.
Each answer agent then selects several review agents through an adaptive social network, simulating the collaborative relationships between agents, to refine its analysis within the visual geo-localization task.
Following this, the framework facilitates open discussions among all answer agents to reach a consensus.
However, as the number of agents grows, managing these discussions can become increasingly chaotic.
Therefore, we introduce a novel dynamic learning strategy to optimize the agent election mechanism and adaptive collaboration network.
By refining both the election and review processes, our framework seeks to discover the most effective communication patterns among agents, thereby improving geo-localization performance through collaborative reasoning while minimizing unnecessary discussions.

In summary, our contributions are demonstrated as follows:

\begin{itemize}[leftmargin=*]
    \item We propose a novel swarm intelligence-based geo-localization framework, \name, 
    that adaptively integrates both the inherent and retrieved knowledge,
    along with the reasoning capabilities of multiple LVLMs, 
    through structured discussions for visual geo-localization tasks.

    \item We introduce a dynamic learning strategy to identify the most effective communication patterns among LVLM agents, enhancing both effectiveness and efficiency.

    \item We conducted experiments on two open-source datasets. To address the issue of numerous images (food, furniture, \etc) in these datasets that could not be localized, we also constructed a new dataset for further evaluation. Extensive experimentation demonstrates that \name\ achieves competitive performance compared to state-of-the-art methods.
\end{itemize}

The remainder of this paper is organized as follows: Section \ref{sec:rw} discusses the related literature. In Section \ref{sec:method}, the proposed framework is introduced. Section \ref{sec:exp} provides the performance evaluation, and Section \ref{sec:cc} concludes the paper.

\section{Related Work}
\label{sec:rw}

\noindent \textbf{Visual Geo-localization}.
Recent research in visual geo-localization, commonly referred to as geo-tagging, primarily focuses on developing image retrieval systems to address this challenge \cite{DBLP:journals/pami/ArandjelovicGTP18,DBLP:journals/fcomp/PaolicelliBMMC22,DBLP:conf/eccv/GeW00L20,jin2017learned,DBLP:conf/iccv/0009LD19,warburg2020mapillary}.
These systems utilize learned embeddings generated by a feature extraction backbone, which includes an aggregation or pooling mechanism \cite{DBLP:conf/icra/PengYZWTW21,DBLP:conf/bmvc/IbrahimiNAW21,DBLP:conf/cvpr/HauslerGXM021,DBLP:journals/pami/RadenovicTC19}.
However, the applicability of these retrieval systems to globally geo-localize landmarks or natural attractions is often limited by the constraints of the available database knowledge and the restrictions imposed by national or regional geo-data protection laws.
Alternatively, some studies treat visual geo-localization as a classification problem \cite{DBLP:conf/pkdd/IzbickiPT19,DBLP:conf/mir/Kordopatis-Zilos21,DBLP:conf/eccv/Muller-BudackPE18,DBLP:conf/eccv/SeoWSH18}.
These approaches posit that two images from the same geographical region, despite depicting different scenes, typically share common semantic features.
Practically, these methods organize the geographical area into discrete cells and categorize the image database accordingly.
This cell-based categorization facilitates scaling the problem globally, provided the number of categories remains manageable.
However, while the number of countries globally remains relatively constant, accurately enumerating cities in real-time at a global scale is challenging due to frequent administrative changes, such as city reorganizations or mergers, which reflect shifts in national policies.
Additionally, in the context of globalization, this strategy has inherent limitations.
The recent advent of LVLMs offers promising compensatory mechanisms for the deficiencies observed in traditional geo-localization methodologies, making the exploration of LVLM-based approaches significantly relevant in current research.

\noindent \textbf{Multi-agent Framework for LLM/LVLMs}.
LLM/LVLM agents have demonstrated the potential to act like human \cite{DBLP:conf/nips/Ouyang0JAWMZASR22,DBLP:journals/corr/abs-2303-12712,DBLP:conf/nips/SchaefferMK23}, and a large number of studies have focused on developing robust architectures for collaborative LLM/LVLM agents \cite{DBLP:conf/acl/Jiang0L23,DBLP:journals/corr/abs-2304-09797,DBLP:conf/nips/ShinnCGNY23,DBLP:journals/corr/abs-2305-14325,DBLP:journals/corr/abs-2310-02170}.
These architectures enable each LLM/LVLM agent that endows with unique capabilities to engage in debates or discussions.
For instance, \cite{DBLP:conf/acl/Jiang0L23} proposes an approach to aggregate multiple LLM/LVLM responses by generating candidate responses from various LLM/LVLM in a single round and employing pairwise ranking to synthesize the most effective response.
While some studies \cite{DBLP:conf/acl/Jiang0L23} utilize a static architecture potentially limiting the performance and generalization of LLM/LVLM, others like \cite{DBLP:journals/corr/abs-2310-02170} have implemented dynamic interaction architectures that adjust according to the query and incorporate user feedback.
Recent advancements also demonstrate the augmentation of LLM/LVLM as autonomous agents capable of utilizing external tools to address challenges in interactive settings.
These techniques include retrieval augmentation \cite{DBLP:journals/corr/abs-2301-12652,DBLP:conf/iclr/YaoZYDSN023,DBLP:journals/jmlr/IzacardLLHPSDJRG23}, mathematical tools \cite{DBLP:conf/nips/SchickDDRLHZCS23,DBLP:conf/iclr/YaoZYDSN023,DBLP:conf/nips/LuPCGCWZG23}, and code interpreters \cite{DBLP:conf/icml/GaoMZ00YCN23,DBLP:journals/corr/abs-2210-12810}.
With these capabilities, LLM/LVLMs are well-suited for various tasks, especially for geo-localization.
However, most LLM/LVLM agent frameworks mandate participation from all agents in at least one interaction round, leading to significant computational overhead.
To address this issue, our framework introduces a dynamic learning strategy electing only a small number of agents to geo-localize different images, which significantly enhances the efficiency of LLM/LVLM agents by reducing unnecessary interactions.

\section{Methodology}
\label{sec:method}

In this section, we first present the overall framework and then introduce each part of \name\ in detail for geo-localization tasks.

\subsection{Model Overview}

\begin{figure*}
\centering
\includegraphics[width=0.9\linewidth]{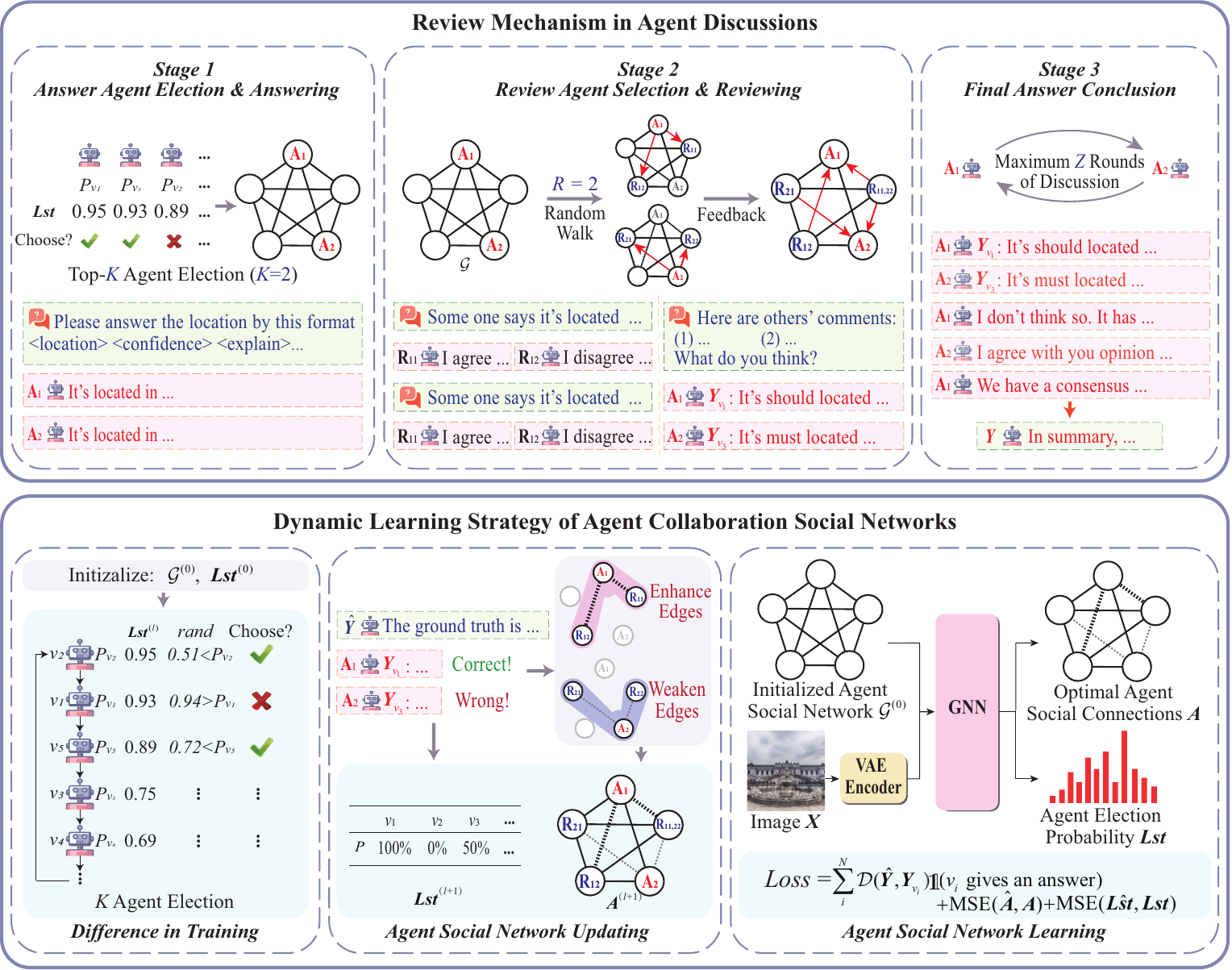}
\caption{
The framework overview of \name.
It contains the process of review mechanism in agent discussions along with a dynamic learning strategy of agent collaboration social networks.
The first part deploys a review mechanism for LVLMs to discuss and share their knowledge anonymously, which could enhance the overall performance of geo-localization tasks.
The second one mainly utilizes the GNN-based learning module to improve efficiency by reducing unnecessary discussions among agents while showing the process of updating the agent collaboration social network during the training process.
}
\label{fig:framework}
\end{figure*}

In this paper, we denote the social network of LVLM agents by $\mathcal{G}$, where $\mathcal{G} = \{ \mathcal{V}, \mathcal{E} \}$.$\mathcal{V}$ stands for the agent set and $\mathcal{E}$ presents the edge set. 
Each agent $v_i \in \mathcal{V}, i \in [N]$ is an LVLM, which is pre-trained by massive vision-language data and can infer the possible location $\boldsymbol{Y}$ of a given image $\boldsymbol{X}$.
Besides, each edge $e_{ij} \in \mathcal{E}, i,j \in [N]$ is the connection weighted by the improvement effect of agent $v_i$ to agent $v_j$ via discussion regarding the geo-localization performance.

As illustrated in Figure \ref{fig:framework}, \name\ contains the process of the review mechanism in agent discussions along with a dynamic learning strategy of agent social networks.
In this framework, the review mechanism in agent discussions is a 3-stage anonymous collaboration approach to allow retrieval-augmented LVLM agents to reach a consensus via discussion. 
In the first stage, for a given image $\boldsymbol{X}$, our framework elects the most suitable $K$ agents as answer agents by agent election probability $\boldsymbol{Lst}$.
In the second stage, these answer agents respectively select $R$ review agents by the adaptive collaboration social network $\boldsymbol{A}$ to refine their answer via discussion.
Finally, our framework facilitates consensus among all agents through open discussion to reach a final answer.
Both $\boldsymbol{Lst}$ and $\boldsymbol{A}$ are analyzed from the given image $\boldsymbol{X}$, allowing our framework to minimize unnecessary discussions, thereby significantly enhancing its efficiency while maintaining its accuracy.
Moreover, the multi-stage discussion facilitates communication among agents, maximizing the integration of their knowledge and reasoning abilities to generate an accurate response $\boldsymbol{Y}$.

To get $\boldsymbol{Lst}$ and $\boldsymbol{A}$, we specifically design a dynamic learning module, which initially deploys the encoder component of a pre-trained image variational autoencoder (VAE) to extract features from the given image $\boldsymbol{X}$.
The extracted features, combined with agent embeddings $\boldsymbol{Emb}$, are employed to determine the suitability of agents \wrt $\boldsymbol{Lst}$ for agent discussions and predict agent collaboration connections $\boldsymbol{A}$ in the geo-localization task.

\subsection{Review Mechanism in Agent Discussions}

% Recognizing the geographic location of unfamiliar images is challenging for an LVLM agent.

LVLM have demonstrated remarkable capabilities in complicated tasks and some pioneering works have further proven that the performances can be further enhanced by ensembling multiple LVLM agents \cite{DBLP:conf/acl/Jiang0L23,DBLP:journals/corr/abs-2304-09797,DBLP:conf/nips/ShinnCGNY23,DBLP:journals/corr/abs-2305-14325,DBLP:journals/corr/abs-2310-02170}.
% However, combining the extensive knowledge and exceptional reasoning capabilities of various LVLMs can yield creative and accurate answers.
Thus, to improve the geo-localization capability of LVLMs, we propose a cooperation framework to effectively integrate the diverse knowledge (including external retrieved information) and reasoning abilities of multiple LVLMs.
Inspired by the fact that community review mechanisms can improve the quality of manuscripts, an iterative 3-stage anonymous reviewing mechanism is proposed for helping agents share knowledge and reasoning capability with each other through their collaboration social network:
i) answer agent election \& answering,
ii) review agent selection \& reviewing,
and iii) final answer conclusion.
% Through multiple rounds of iterations, this process can adaptively discover the most appropriate agent to start a discussion on this visual geo-localization task and give the correct result.

\noindent \textbf{Stage 1: Answer Agent Election \& Answering}

Initially, we select $K$ agents with the highest agent election probabilities~$\boldsymbol{Lst}$ as answer agents and let them geo-localize independently as the preliminary step for further discussion.
By initiating the discussion with a limited number of agents, we aim to reduce potential chaos and maintain the efficiency of our framework as the number of participating agents increases. 
% We hope by just choosing a few agents to initiate discussion, we can eliminate the chaos in discussion among agents coming from the increasing of the number of agents and keep our framework efficient.

% ,  as the number of agents increases, 
% % giving an image to all LVLM agents and then 
% letting them freely discuss among themselves would make the whole geo-localization framework become chaos and inefficient.
% Thus, choosing suitable agents for the initial location analysis is conducive to improving the efficiency of the framework.
% Specifically, we first sort the agent probability scalars in descending order by agent probability value: $\boldsymbol{Lst}_{\text{sorted}} = \mathrm{sort}([P_{v_1}, P_{v_2}, \cdots, P_{v_N}])$.
% Then, we sequentially select top-$K$ agents
% as illustrated at the top of Stage 1 in Figure \ref{fig:framework}.

After the answer agents are elected, we send the image $\boldsymbol{X}$ to all answer agents and let them give the primary analysis.
% , as shown at the bottom of Stage 1 in Figure \ref{fig:framework}.
Each answer must contain three parts: one location (city, country, and so on), one confidence (a percentage number), and a detailed explanation.
% This structure is illustrated at the bottom of Stage 1 in Figure \ref{fig:framework}.
Besides, if an agent is unable to interpret the given image, it is permitted to utilize a combination of network search and chain-of-thought reasoning \cite{DBLP:conf/nips/Wei0SBIXCLZ22} to gather additional relevant information.

\noindent \textbf{Stage 2: Review Agent Selection \& Reviewing}

In this stage, for each answer agent, we choose ${R}$ retrieval-augmented review agents by performing a transfer-probability-based random walk on the agent collaboration social network $\mathcal{G}$ for answer reviewing. %, as shown at the left part of Stage 2 in Figure \ref{fig:framework}.
The transfer probability $p(v_i,v_j)$ from node $v_i$ to node $v_j$ can be calculated as follows:

\begin{equation}
\label{eq:tp}
p(v_i,v_j) = 
\begin{cases}
\frac{\boldsymbol{A}_{ij}}{\sum_{k \in \mathcal{N}(v_i)} \boldsymbol{A}_{ik}}, &\text{if}\ e_{ij} \in \mathcal{E}, \\
0, &\text{otherwise},
\end{cases}
\end{equation}
where $\mathcal{N}(v_i)$ is the 1-hop neighbor node set of node $v_i$. 

For each selected review agent, it reviews the results as well as the explanations generated by the corresponding answer agent and gives its own comments. 
After that, 
% as illustrated at the right part of Stage 2 in Figure \ref{fig:framework}, 
each answer agent would summarize their preliminary analysis and the feedback from all of its review agents to get the final answer, which must include three parts as well: one location, one confidence, and an explanation.

% as shown at the bottom of stage 2 in Figure \ref{fig:framework}.

% \input{draw/Review}

\noindent \textbf{Stage 3: Final Answer Conclusion}

In the previous stage, each answer agent produces a refined result based on feedback.
When $K>1$ in Stage 1, the proposed framework generates multiple independent results, which may not be consistent.
However, we aim to provide a definitive answer rather than multiple options for people to choose from.
To address this, we allow up to $Z$ rounds of free discussion among those answer agents to reach a unified answer:

First, we maintain a global dialog history list, $diag$, recording all replies agents respond.
In addition, discussions are executed asynchronously, which means that any answer agent can always reply based on the latest $diag$, and replies would be added to the end of $diag$ as soon as they are posted.
Each answer agent is allowed to speak only once in each discussion round, and after $Z$ rounds of free discussion, we determine the final result using a minority-majority approach,
\ie we choose the reply with the most agreement as the final conclusion.
If all agents reach a consensus (or when the conclusion of the majority of agents is clearly dominant), we early stop this stage and adopt the consensus answer as the final answer.
If none of any consensus is reached, we only select the reply of the first answer agent elected from Stage 1 as the final result.

\subsection{Dynamic Learning Strategy of Agent Collaboration Social Networks}
\label{subsec:asnl}

In our framework, choosing the appropriate answer agents and review agents for knowledge sharing and discussion is vital to its effectiveness and efficiency.
% However, it is tough to select a suitable agent without human intervention. 
Therefore, we propose a dynamic learning strategy to optimize them.
Specifically, for each training sample, \ie a geo-tagged image, we would first estimate the optimal answer agent election probability $\boldsymbol{\hat{Lst}}$ and the optimal collaboration social network of agent $\mathcal{\hat{G}}$ by its actual location.
Then we train an attention-based graph neural network, which aims to predict $\boldsymbol{Lst}$ and $\mathcal{G}$, by such estimated ground truth.
% for adaptive learning of election mechanism and agent social relationships.
% Therefore, we propose a "small" dynamic agent social network learning module for adaptive learning of agent social relationships in locating the given landmark images.

To estimate the optimal $\boldsymbol{\hat{Lst}}$ and $\boldsymbol{\hat{A}}$ for agents to geo-localize image $\boldsymbol{X}$, 
we first initialize the agent social network $\mathcal{G}^{(0)}$ by a fully connected graph with the agent set $\mathcal{V}$.
Besides, we initialize the agent election probability $\boldsymbol{Lst}^{(0)} = [0.5, 0.5, \cdots]$, with all agents having $50\%$ probability of being chose as answer agents.

Then, we iteratively conduct our 3-stage discussion framework to get the prediction answer.
% During the training period, agents are allowed $L$ rounds of discussions.
$\boldsymbol{Lst}^{(l)}$ and $\mathcal{G}^{(l)}$ is updated at the end of each round $l \in L$ by comparing the answers $\boldsymbol{Y}_{v_i}^{(l)}$ from each answer agent with the ground truth $\hat{\boldsymbol{Y}}$.

After $L$ rounds of agent discussions, the updated agent election probability for an image $\boldsymbol{X}$, $\boldsymbol{\hat{Lst}} := \boldsymbol{Lst}^{(L)}(\boldsymbol{{X}}) = [P_{v_1}^{(L)}, P_{v_2}^{(L)}, \cdots, P_{v_N}^{(L)}]$, determines whether an agent $v_i$ gives the correct/wrong answers $\boldsymbol{Y}_{vi}^{(L)}$ by comparing it with the ground truth $\hat{\boldsymbol{Y}}$.
Here, the definition of $P_{v_i}^{(l)}$ of agent $v_i$ at round $l$ is as follows:

\begin{equation}
\label{eq:agent_answer}
P_{v_i}^{(l)} :=
\begin{cases} 
0, &\text{if}\ \mathcal{D}(\hat{\boldsymbol{Y}}, \boldsymbol{Y}_{v_i}^{(l)}) > th, \\
1, &\text{if}\ \mathcal{D}(\hat{\boldsymbol{Y}}, \boldsymbol{Y}_{v_i}^{(l)}) \leq th, \\
\frac{1}{2}, &\text{if }v_i\text{ did not participate in the discussion},
\end{cases}
\end{equation}
where $th$ is a pre-defined threshold for determining whether the predicted location is close enough to the actual location.
In the distance function $\mathcal{D}(\cdot)$, we first deploy geocoding to convert natural language into location intervals in a Web Mercator coordinate system (WGS84) by utilizing OSM APIs, and then compute the shortest distance between two two location intervals.

Please note that, rather than electing the top-$K$ answer agents in each round, we choose each agent with probability $P_{v_i}$ during the training period to ensure that every agent has the opportunity to participate in the discussion for more accurate estimation, as shown at the left part of the dynamic learning strategy module of agent collaboration social networks in Figure \ref{fig:framework}.

% in some of the total $L$ rounds 

In addition, the agent collaboration social network would also be updated by comparing the actual location with the generated answer of each answer agent at the same time. 
For $l$-th round, we strengthen the link between the correctly answered agent and the corresponding review agents while weakening the link between the incorrectly answered agent and the corresponding review agents:

\begin{equation}
\label{eq:link_update}
\boldsymbol{\hat{A}}_{ij} := \boldsymbol{A}_{ij}^{(l)}(\boldsymbol{X}) =
\begin{cases}
 \frac{tt + 1}{2 tt} \boldsymbol{A}_{ij}^{(l-1)}(\boldsymbol{X}), &\text{if }v_i\text{ answers correctly}, \\
\frac{2 tt - 1}{2 tt} \boldsymbol{A}_{ij}^{(l-1)}(\boldsymbol{X}), &\text{if }v_i\text{ answers incorrectly}, \\
\end{cases}
\end{equation}
where $\boldsymbol{A}_{ij}^{(l-1)}(\boldsymbol{X})$ is the weight of the connection between answer agent $v_i$ and review agent $v_j$ at round $l-1$ when geo-locating image $\boldsymbol{X}$,  $\boldsymbol{A}^{(0)}_{ij}(\boldsymbol{X}) = 1, i \neq j, \boldsymbol{A}^{(0)}_{ii}(\boldsymbol{X}) = 0, i,j \in [N]$, $tt$ is the number of consecutive times an agent has answered correctly, which is used to attenuate the connection weights when updating them, preventing the performance of an agent on a certain portion of the continuous dataset from interfering with the model's evaluation of the current agent's performance on the entire dataset.

% In the dynamic agent social network learning module, 
Then, we try to learn an attention-based graph neural network to predict the corresponding optimal agent election probability $\boldsymbol{Lst} = h(\boldsymbol{X}, \mathcal{G} | \Theta)$ and the optimal agent collaboration connections $\boldsymbol{A} = f(\boldsymbol{X}, \mathcal{V} | \Theta)$: 
% parameterized by $\Theta$ to engrave optimal communication circles for agents where two agents $v_i$ and $v_j$ could have a conversation at probability ${\boldsymbol{A}'}_{ij}$, and it is also utilized to choose $K$ suitable agents by predicting the agent probability scalar $\boldsymbol{Lst}' = h(\boldsymbol{X}, \mathcal{G} | \Theta)$ for answering and concluding the final results.
% Here, we utilize an attention-based graph neural network to predict $\boldsymbol{A}'$ and $\boldsymbol{Lst}'$:

\begin{equation}
\label{eq:gann}
\centering
\begin{aligned}
\boldsymbol{A} &= \mathrm{Att}_{\text{GNN}}(\boldsymbol{Fea}, \boldsymbol{Fea}, \boldsymbol{1}) \\ &= \mathrm{softmax}\left( \frac{\boldsymbol{Fea} \cdot \boldsymbol{Fea}^{\top}}{\sqrt{d_k}} \right) 
\boldsymbol{1}, \\
% \mathrm{Norm}(2\boldsymbol{A}^{(0)}), \\
\boldsymbol{Lst} &=\sigma'\left(\mathrm{Linear}\left(\mathrm{Flatten}\left(\sigma\left(\boldsymbol{A} \cdot \boldsymbol{Fea} \cdot \boldsymbol{W}\right)\right)\right)\right), \\
\boldsymbol{Fea} &= \mathrm{Linear}\left(\boldsymbol{Emb}+\mathrm{VAE_{Enc}}(\boldsymbol{X})\right), \\
% \boldsymbol{X} &= \mathrm{VAE_{Dec}}\left(\sigma\left(\mathrm{VAE_{Enc}}(\boldsymbol{X}) \right)\right)
\end{aligned}
\end{equation}
where $\boldsymbol{W}, \boldsymbol{Emb} \in \Theta$ are two learnable parameters, $\boldsymbol{Emb} := [\boldsymbol{Emb}_{v_1}, $ $\boldsymbol{Emb}_{v_2}, \cdots]^{\top}$ is the agent embedding and $\boldsymbol{W}$ is the weight matrix, $\sigma(\cdot)$ is the LeakyReLU function, $\sigma'(\cdot)$ is the Sigmoid function, $\mathrm{VAE_{Enc}}(\cdot)$ is the encoder of the image VAE that compresses and maps the image data into the latent space. It is used to align the image features with the agent embedding, and $d_k$ is the dimension of the $\boldsymbol{Fea}$.
% In addition, to avoid the attention mechanism resulting each element in  $\boldsymbol{A}'$ is always less than or equal to the one in $\boldsymbol{A}$, we scale and normalize the $\boldsymbol{A}$ when applying the attention mechanism, where $\mathrm{Norm(\cdot)}$ is the min-max normalization that scaled the origin adjacency matrix of the agent social network $\mathcal{G}$.
% By leveraging the strengths of different LLM agents in capturing different details, agents under a geo-localization-specific social network could communicate freely with others and summarize the optimal final result $\boldsymbol{Y}'$ after $L$ rounds of discussions among LVLM agents. 
Our learning target can be formalized as:

\begin{equation}
\label{eq_final_loss}
\centering
\begin{aligned}
\arg \min_{\Theta}  \sum_
{i}^{N}\mathcal{D}(\boldsymbol{\hat{Y}}, \boldsymbol{Y}_{v_{i}}) &\mathbbm{1}(v_{i} \text{ gives an answer}) \\
& + \mathrm{MSE}(\boldsymbol{\hat{Lst}}, \boldsymbol{Lst}) + \mathrm{MSE}(\boldsymbol{\hat{A}}, \boldsymbol{A}),
\end{aligned}
\end{equation}
where $\mathcal{D}(\cdot)$ denotes the distance between the places an LVLM agent answered and the ground truth, $\mathbbm{1}(\cdot)$ is the indicator function, $\boldsymbol{Y}_{v_i} := \boldsymbol{Y}_{v_i}^{(L)} = g_{v_i}(\boldsymbol{X}, \boldsymbol{Y}_{v_j}^{(L-1)})$, $g_{v_i}(\cdot)$ represent the LVLM agent $v_i$ with fixed parameters and $\boldsymbol{Y}_{v_i}^{(0)}=g_{v_i}(\boldsymbol{X})$ is the answer that LVLM agent $v_i$ generates at the initial stage of discussion.

\section{Experiments}
\label{sec:exp}

To evaluate the performance of our framework, we conducted experiments on the real-world dataset that was gathered from the Internet to answer the following research questions:

\noindent $\bullet$ \textbf{RQ1}: Can the proposed framework, \name, outperform state-of-the-art methods in open-ended geo-localization tasks?

\noindent $\bullet$ \textbf{RQ2}: Are LVLMs with diverse knowledge and reasoning abilities more suitable for building a collaborative social network of agents?
    
\noindent $\bullet$ \textbf{RQ3}: How efficient is \name compared to other baselines?

\noindent $\bullet$ \textbf{RQ4}: How does the setting of different hyperparameters affect the performance of \name?

\subsection{Experiment Setup}

\noindent \textbf{Datasets}.
In this paper, we first evaluate the proposed geo-localization framework, \name,
on the two open-source datasets: IM2GPS3K\footnote{http://www.mediafire.com/file/7ht7sn78q27o9we/im2gps3ktest.zip} and YFCC4K\footnote{http://www.mediafire.com/file/3og8y3o6c9de3ye/yfcc4k.zip}:

The IM2GPS3K dataset is a widely used benchmark for visual geo-localization. It consists of 3,000 images from the Flickr photo-sharing platform, which are tagged with precise geographical coordinates. The dataset covers a diverse range of locations globally, including urban, rural, and natural environments, making it an ideal testbed for assessing the generalization capabilities of geo-localization models. The images in IM2GPS3K are drawn from various categories, such as landscapes, cityscapes, and landmarks, providing a challenging and diverse set of visual cues that the models must recognize to predict the geographical location accurately.

    The YFCC4K dataset is a subset of the larger Yahoo Flickr Creative Commons 100 Million (YFCC100M) dataset, which contains over 100 million Flickr images. The YFCC4K subset includes 4,536 images with corresponding geographical tags, selected to represent a broad geographic distribution and visual diversity. Like IM2GPS3K, YFCC4K includes a variety of image types, such as natural landscapes, urban settings, and iconic landmarks. This diversity allows for comprehensive evaluation of geo-localization models across different environments and scales, from global-level to city-level localization tasks.

Noting that the labels in the above two datasets are latitude-and-longitude-based GPS points rather than natural language, we use geo-reverse encoding technology to map GPS points into detailed and structured addresses for evaluation.

In addition,
We have newly constructed a geo-localization dataset named GeoGlobe.
It contains a variety of man-made landmarks or natural attractions from nearly 150 countries with different cultural and regional styles.
The diversity and richness of GeoGlobe allow us to evaluate the performance of different models more accurately.
% \footnote{\href{https://www.kaggle.com/datasets/o0o0oo/open-source-geolocalization-dataset-for-paper-xxx}{\re{https://www.kaggle.com/datasets/o0o0oo/open-source-geolocalization-dataset-for-paper-xxx} }}
More details can be found in Appendix \ref{app:dataset}.

\noindent \textbf{Baselines}.
In this work, we mainly compare the proposed framework with different geolocation retrieval systems and deep learning methods: including NetVLAD \cite{DBLP:journals/pami/ArandjelovicGTP18}, GeM \cite{DBLP:journals/pami/RadenovicTC19}, CosPlace \cite{DBLP:conf/cvpr/BertonMC22}, Translocator \cite{pramanick2022world}, and GeoCLIP \cite{vivanco2024geoclip}.
For image retrieval systems, we set the whole training dataset as the geo-tagged image database and only query images in the test dataset for systems to answer.

% GeoSpy\footnote{\href{https://geospy.ai/}{https://geospy.ai/}}

\noindent \textbf{Implemention Details}.
We select both open-source and close-source LVLMs with different scales pretrained by various datasets as agents in the proposed framework.
As for the open-source LVLMs, we utilize several open-source fine-tuned LVLMs:
Infi-MM\footnote{\href{https://huggingface.co/Infi-MM/infimm-zephyr}{https://huggingface.co/Infi-MM/infimm-zephyr}},
Qwen-VL \footnote{\href{https://huggingface.co/Qwen/Qwen-VL}{https://huggingface.co/Qwen/Qwen-VL}},
vip--llava--7b\&13b\footnote{\href{https://huggingface.co/llava-hf/}{https://huggingface.co/llava-hf/vip-llava-xxx}},
llava--1.5--7b--base\&mistral\&vicuna\footnote{\href{https://huggingface.co/llava-hf/}{https://huggingface.co/llava-hf/llava-1.5-xxx}},
llava--1.6--7b\&13b\&34b--mistral\&vicuna\footnote{\href{https://huggingface.co/liuhaotian/}{https://huggingface.co/liuhaotian/llava-v1.6-xxx}},
CogVLM\footnote{\href{https://github.com/THUDM/CogVLM}{https://github.com/THUDM/CogVLM}}.
As for the closed-source LVLMs, we chose the models provided by three of the most famous companies in the world:
Claude--3--opus\footnote{\href{https://anthropic.com/}{https://anthropic.com/}},
GPT--4o--mini\footnote{\href{https://openai.com/}{https://openai.com/}},
and Gemini--1.5--pro \footnote{\href{https://gemini.google.com/}{https://gemini.google.com/}}.
Besides, 99\% of images from each dataset are randomly chosen as training samples.
For the open-world geo-localization problem, we constructed the test dataset that is entirely independent of each training dataset. Additionally, approximately 50\%-60\% of the test samples consist of images from distinct locations with no overlap with the training data.
More details about the deployment of \name\ and the settings of related hyperparameters can be found in Appendix \ref{app:implementation}.

% \re{single agent}

% \re{retrival method: 99\% train, 1\% test}

% \re{agent framework}

\noindent \textbf{Evaluation Metrics}.
We use \emph{Accuracy} ($Acc$) to evaluate the performance:
$Acc = \frac{N_{\text{correct}}}{N_{\text{total}}}$,
where $N_{\text{correct}}$ is the number of samples that the proposed framework correctly geo-localizes, and $N_{\text{total}}$ refers to the total number of testing samples.

In this paper, we first geo-encode the answers with the ground truth, \ie we transform the addresses described through natural language into latitude-longitude coordinates.
Then, we calculate the distance between the two coordinates.
When the distance between the two coordinates is less than $th=50 km$ (city-level), we consider the answer of the framework to be correct.

\subsection{Performance Comparison}

% \begin{wraptable}{r}{0.53\textwidth}
\begin{table}
\centering
\captionof{table}{Comparison with baselines.}
\label{tab:exp_baselines}
\renewcommand\tabcolsep{2.5pt}
\begin{tabular}{ccccc} % 开始一个表格
\hline
\bfseries{Model} & \bfseries{IM2GPS3K} & \bfseries{YFCC4K} & \makecell[c]{\bfseries{GeoGlobe}\\\bfseries{(Natural)}} & \makecell[c]{\bfseries{GeoGlobe}\\\bfseries{(ManMade)}}\\ \hline
NetVLAD & 16.6303 & 7.4876 & 26.5134 & 28.9955 \\
GeM & 14.4907 & 6.5243 & 23.1022 & 25.4175 \\
CosPlace & 17.6686 & 7.9551 & 28.1688 & 30.2782 \\
Translocator & 31.0978 & 13.4039 & 26.1776 & 34.1971 \\
GeoCLIP & 34.4728 & 15.1719 & 38.2519 & 45.9174  \\ \hline
\textbf{\name} & \textbf{47.7666} & \textbf{21.5064} & \textbf{76.1535} & \textbf{85.4548} \\ \hline
\end{tabular}

\small{
Bold indicates the statistically significant improvements \\ (\ie two-sided t-test with $p < 0.05$) over the best baseline.
}
% \end{wraptable}
\end{table}

% \begin{table*}[!ht]
% \caption{Results of different single LVLM baselines.}
% \label{tab:baselines}
% \centering
% % \renewcommand\tabcolsep{2.5pt}
% \csvreader[
%   tabular=cccccccc,
%   table head=\hline
  
%   \specialrule{0em}{1pt}{1pt}

%   \multicolumn{1}{c}{} & \multicolumn{3}{c}{\bfseries{Without Web Searching}} & \multicolumn{3}{c}{\bfseries{With Web Searching}}
  
%   \\ \specialrule{0em}{1pt}{1pt}

%   \multicolumn{1}{c}{} & \multicolumn{1}{c}{\bfseries{Natural}} & \multicolumn{1}{c}{\bfseries{ManMade}} & \multicolumn{1}{c}{\bfseries{Overall}} & \multicolumn{1}{c}{\bfseries{Natural}} & \multicolumn{1}{c}{\bfseries{ManMade}} & \multicolumn{1}{c}{\bfseries{Overall}}
  
%   \\ \specialrule{0em}{1pt}{1pt}

%   \hline,
%   late after line={\\},
%   late after last line= \\
%   \hline % horizontal line at the end of the table
% ]{data/baselines.txt}{}{

% \ifnumequal{\thecsvrow}{11}{ \hline }{}
% \ifnumequal{\thecsvrow}{14}{ \hline }{}
% \csvlinetotablerow
% }

% \small{
% Bold indicates the statistically significant improvements \\ (\ie two-sided t-test with $p < 0.05$) over the best baseline.
% }

% \end{table*}

Table \ref{tab:exp_baselines} presents a comparison between our proposed framework and all baseline approaches. Our framework consistently outperforms all other methods.
This superior performance can be attributed to the limitations of traditional image retrieval techniques, which rely heavily on rich geo-tagged image databases and exhibit constrained reasoning capabilities.
In contrast, our method can effectively analyze and integrate results retrieved from the Internet, enabling it to calculate more accurate geo-localization outcomes.
Moreover, over half of the images in our test dataset are new and localized in areas completely different from those in the training dataset.
This underscores the shortcomings of conventional database-based retrieval systems, particularly due to the inherent limitations of geo-tagged image databases, and demonstrates the effectiveness of our framework in addressing open-world geo-localization tasks.  

It is also worth noting that the YFCC4K and IM2GPS3K datasets do not apply artificial filtering to images, resulting in ambiguous content with minimal geographical cues, such as food photos and portraits. Comparing the model performance across different datasets, we observe that it performs best on the GeoGlobe dataset.

\subsection{Ablation Study}
\label{subsec:ablation}

\begin{table}[!ht]
\caption{Results of different single LVLM baselines.}
\label{tab:single_agents}
\centering
\renewcommand\tabcolsep{1pt}
\begin{filecontents*}{data/single_agents.txt}
model,img2gps,yfcc,nature,human
Infi-MM,14.7000,6.5256,19.2020,21.4145
Qwen-VL,32.4667,14.3959,42.3940,37.4556
vip-llava-13b,15.7667,6.9885,20.6983,15.4089
vip-llava-7b,38.7000,17.8792,31.9202,56.4994
llava-1.5-7b,30.2000,13.8889,27.1820,47.2145
llava-1.6-7b-mistral,3.1000,1.6314,0.7481,2.1731
llava-1.6-7b-vicuna,9.4667,4.4092,6.9825,15.8831
llava-1.6-13b,17.8333,8.3333,12.2195,28.2497
llava-1.6-34b,44.2667,20.8113,52.8678,77.2027
CogVLM,6.7333,3.0644,7.7307,10.3516
Claude-3-opus,23.8333,12.3457,33.1671,40.6954
GPT-4o-mini,45.0000,18.3422,62.0948,84.5911
Gemini-1.5-pro,47.3667,19.9956,62.3441,82.8131
\textbf{\name},\textbf{47.7667},\textbf{21.5168},\textbf{76.0599},\textbf{85.4603}

\end{filecontents*}
\csvreader[
  tabular=ccccc,
  table head=\hline
  \specialrule{0em}{1pt}{1pt}
  \multirow{2}{*}{\bfseries{Model}} & \multirow{2}{*}{\bfseries{IM2GPS3K}} & \multirow{2}{*}{\bfseries{YFCC4K}} & \bfseries{GeoGlobe} & \bfseries{GeoGlobe} \\ & & & \bfseries{(Natural)} & \bfseries{(ManMade)}
  \\ \specialrule{0em}{1pt}{1pt}
  \hline,
  late after line={\\},
  late after last line=\\
  \hline % horizontal line at the end of the table
]{data/single_agents.txt}{}{
\ifnumequal{\thecsvrow}{11}{ \hline }{}
\ifnumequal{\thecsvrow}{14}{ \hline }{}
\csvlinetotablerow
}

\small{
Bold indicates the statistically significant improvements \\ (\ie two-sided t-test with $p < 0.05$) over the best baseline.
}

\end{table}
\noindent \textbf{Effect of LVLM Discussion}. 
In this section, we aim to verify that the observed performance improvements stem from the collaborative discussion among multiple LVLM agents rather than from a single LVLM agent.
To achieve this, we employ the chain-of-thought (CoT) method \cite{DBLP:conf/nips/Wei0SBIXCLZ22} to evaluate each agent individually within our framework, and the results are presented in Table \ref{tab:single_agents}. 
We can find that \name\ outperforms any individual LVLM across all datasets, including the most advanced closed-source models such as GPT-4o-mini and Gemini-1.5-pro. Furthermore, the varying performances of different single LVLMs on different datasets highlight that each agent possesses distinct knowledge and reasoning capabilities.
Our proposed framework facilitates effective information exchange among agents, thereby enhancing the reasoning abilities of the models for diverse geo-localization tasks.

\noindent \textbf{Power of Internet-enabled  Information Retrieval}.
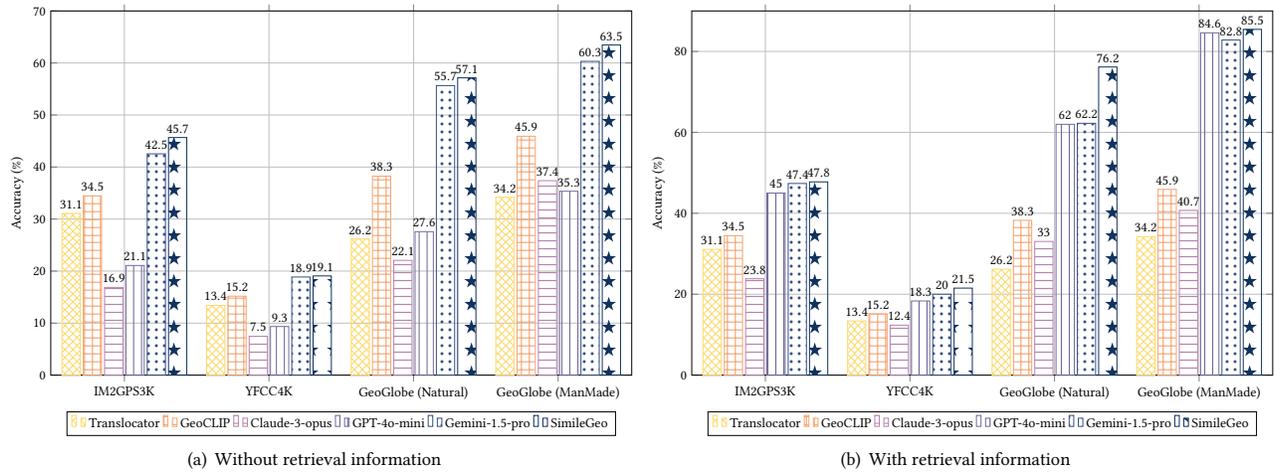
\begin{figure*}[h!]
\centering
\subfigure[Without retrieval information]{
\label{fig:retrieval_a}
\resizebox{0.47\linewidth}{!}{
\begin{tikzpicture}
    \begin{axis}[
        ybar,
        grid=major,
        bar width=12pt,
        width=15cm, height=10cm,
        enlarge x limits={abs=1.7cm},
        legend style={at={(0.5,-0.1)}, anchor=north, legend columns=-1},
        symbolic x coords={IM2GPS3K, YFCC4K, GeoGlobe (Natural), GeoGlobe (ManMade)},
        xtick=data,
        ylabel={Accuracy (\%)},
        nodes near coords,
        every node near coord/.append style={/pgf/number format/.cd, fixed, precision=1}, % 保留1位小数
        ymin=0,
        ymax=70,
        xticklabel style={rotate=0, anchor=north},
        ]
        
        % Data for Translocator
        \addplot [pattern=crosshatch, pattern color=mycolor66, draw=mycolor66] coordinates {(IM2GPS3K,31.0978) (YFCC4K,13.4039) (GeoGlobe (Natural),26.1776) (GeoGlobe (ManMade),34.1971)};
        % Data for GeoCLIP
        \addplot [pattern=grid, pattern color=mycolor65, draw=mycolor65] coordinates {(IM2GPS3K,34.4728) (YFCC4K,15.1719) (GeoGlobe (Natural),38.2519) (GeoGlobe (ManMade),45.9174)};
        % Data for Claude-3-opus
        \addplot [pattern=horizontal lines, pattern color=mycolor64, draw=mycolor64] coordinates {(IM2GPS3K,16.8604) (YFCC4K,7.4768) (GeoGlobe (Natural),22.0600) (GeoGlobe (ManMade),37.3801)};
        % Data for GPT-4o-mini
        \addplot [pattern=vertical lines, pattern color=mycolor63, draw=mycolor63] coordinates {(IM2GPS3K,21.0775) (YFCC4K,9.3469) (GeoGlobe (Natural),27.5776) (GeoGlobe (ManMade),35.3443)};
        % Data for Gemini-1.5-pro
        \addplot [pattern=dots, pattern color=mycolor62, draw=mycolor62] coordinates {(IM2GPS3K,42.5348) (YFCC4K,18.8621) (GeoGlobe (Natural),55.6522) (GeoGlobe (ManMade),60.3107)};
        % Data for SimileGeo
        \addplot [pattern=fivepointed stars, pattern color=mycolor61, draw=mycolor61] coordinates {(IM2GPS3K,45.6691) (YFCC4K,19.0714) (GeoGlobe (Natural),57.1476) (GeoGlobe (ManMade),63.4532)};
        
        \legend{Translocator, GeoCLIP, Claude-3-opus, GPT-4o-mini, Gemini-1.5-pro, SimileGeo}
    \end{axis}
\end{tikzpicture}}}
\subfigure[With retrieval information]{
\label{fig:retrieval_b}
\resizebox{0.47\linewidth}{!}{
\begin{tikzpicture}
    \begin{axis}[
        ybar,
        grid=major,
        bar width=12pt,
        width=15cm, height=10cm,
        enlarge x limits={abs=1.7cm},
        legend style={at={(0.5,-0.1)}, anchor=north, legend columns=-1},
        symbolic x coords={IM2GPS3K, YFCC4K, GeoGlobe (Natural), GeoGlobe (ManMade)},
        xtick=data,
        ylabel={Accuracy (\%)},
        nodes near coords,
        every node near coord/.append style={/pgf/number format/.cd, fixed, precision=1}, % 保留1位小数
        ymin=0,
        ymax=90,
        xticklabel style={rotate=0, anchor=north},
        ]
        
        % Data for Translocator
        \addplot [pattern=crosshatch, pattern color=mycolor66, draw=mycolor66] coordinates {(IM2GPS3K,31.0978) (YFCC4K,13.4039) (GeoGlobe (Natural),26.1776) (GeoGlobe (ManMade),34.1971)};
        % Data for GeoCLIP
        \addplot [pattern=grid, pattern color=mycolor65, draw=mycolor65] coordinates {(IM2GPS3K,34.4728) (YFCC4K,15.1719) (GeoGlobe (Natural),38.2519) (GeoGlobe (ManMade),45.9174)};
        % Data for Claude-3-opus
        \addplot [pattern=horizontal lines, pattern color=mycolor64, draw=mycolor64] coordinates {(IM2GPS3K,23.8245) (YFCC4K,12.3522) (GeoGlobe (Natural),33.0435) (GeoGlobe (ManMade),40.7125)};
        % Data for GPT-4o-mini
        \addplot [pattern=vertical lines, pattern color=mycolor63, draw=mycolor63] coordinates {(IM2GPS3K,44.9928) (YFCC4K,18.3322) (GeoGlobe (Natural),61.9876) (GeoGlobe (ManMade),84.6028)};
        % Data for Gemini-1.5-pro
        \addplot [pattern=dots, pattern color=mycolor62, draw=mycolor62] coordinates {(IM2GPS3K,47.3686) (YFCC4K,20.0035) (GeoGlobe (Natural),62.2360) (GeoGlobe (ManMade),82.8206)};
        % Data for SimileGeo
        \addplot [pattern=fivepointed stars, pattern color=mycolor61, draw=mycolor61] coordinates {(IM2GPS3K,47.7666) (YFCC4K,21.5064) (GeoGlobe (Natural),76.1535) (GeoGlobe (ManMade),85.4548)};
        
        \legend{Translocator, GeoCLIP, Claude-3-opus, GPT-4o-mini, Gemini-1.5-pro, SimileGeo}
    \end{axis}
\end{tikzpicture}}}
\caption{The impact of Internet-enabled information retrieval.}
\label{fig:retrieval}
\end{figure*}

To demonstrate the benefits of allowing every LVLM agent to retrieve information from the Internet for enhancing the knowledge of the model, as well as the potential for multiple LVLMs to use the retrieved data for further reasoning, we designed an experiment as shown in Figure \ref{fig:retrieval}.
Figure~\ref{fig:retrieval_a} illustrates the results without any retrieval, while Figure~\ref{fig:retrieval_b} displays the results when models incorporate the retrieved information.
We compare the proposed model with two types of methods: image similarity-based retrieval approaches and advanced closed-source large models. The results indicate that image similarity-based methods, such as Translocator and GeoCLIP, struggle to effectively leverage Internet-retrieved information, leading to minimal changes in accuracy. In contrast, closed-source large models show moderate performance improvements when aided by additional retrieval data. Notably, our method consistently outperforms all baselines across datasets, with a significant improvement of over 20\% in the GeoGlobe (natural) dataset for \name. This demonstrates that our proposed framework, with its robust reasoning capabilities, can fully exploit retrieved information to enhance reasoning and model accuracy.

\noindent \textbf{Different LVLM Agent Structures}.
\begin{table*}[ht]
\centering
\caption{Results of different agent frameworks testing on GeoGlobe (ManMade).}
\label{tab:exp_structure}
\newcolumntype{C}[1]{>{\centering\arraybackslash}m{#1}}
\begin{tabular}{C{1.5cm}C{1.5cm}C{1.5cm}C{1.5cm}C{1.5cm}C{1.5cm}C{1.5cm}}
\hline
\bfseries{Framework} & \bfseries{LLM-Blender} & \bfseries{PHP} & \bfseries{Reflexion} & \bfseries{LLM Debate} & \bfseries{DyLAN} & \textbf{\name} \\ \hline
Sturcture & \includegraphics[width=1.5cm]{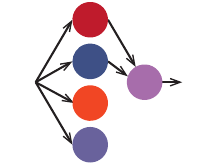} & \includegraphics[width=1.5cm]{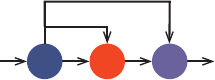} & \includegraphics[width=1.35cm]{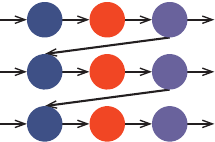} & \includegraphics[width=1.35cm]{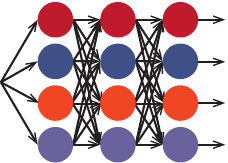} & \includegraphics[width=1.5cm]{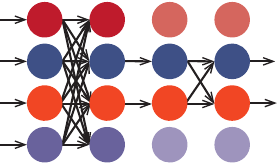} & \includegraphics[width=1.4cm]{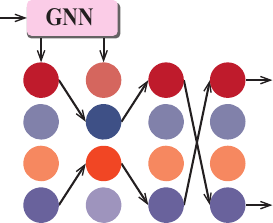} \\ \hline
$Acc (\%)$ $\uparrow$ & 75.3352\% & 82.3592\% & 84.1963\% & 76.9987\% & 84.8412\% & \textbf{85.4548\%} \\ 
$Tks$ $\downarrow$ & 23,662 & 154,520 & 109,524 & 260,756 & 159,320 & \textbf{17,145} \\ \hline
\end{tabular}

\small{
'$Acc$' stands for the accuracy of the framework; \\
'$Tks$' means the average tokens a framework costs per query (including image tokens).
}

\end{table*}
We experiment with multi-agent collaborative frameworks with different structures, including LLM-Blender \cite{DBLP:conf/acl/Jiang0L23}, PHP \cite{DBLP:journals/corr/abs-2304-09797}, Reflexion \cite{DBLP:conf/nips/ShinnCGNY23}, LLM Debate \cite{DBLP:journals/corr/abs-2305-14325}, and DyLAN \cite{DBLP:journals/corr/abs-2310-02170}.
The comparative results across various LVLM agent frameworks are presented in Table \ref{tab:exp_structure}.
It is evident that the majority of LVLM agent frameworks surpass individual LVLMs in terms of geo-localization accuracy in Table 
\ref{tab:single_agents}.
This improvement can primarily be attributed to the ability to integrate knowledge from multiple LVLM agents, thereby enhancing the overall precision of these frameworks.
However, LLM-Blender and LLM Debate exhibit lower accuracy due to statements of some agents misleading others during discussions, which impedes the generation of correct outcomes.
Our framework, \name, guarantees the highest accuracy while being able to accomplish the geo-localization task with the lowest token costs.
The average number of tokens our framework spent per query is 17,145, and it is less than the computational overhead of LLM-Blender (23,662), which has the simplest agent framework structure but the lowest accuracy among all baselines.
This is mainly due to a 'small' GNN-based dynamic learning model being deployed for agent selection stages and significantly reducing unnecessary discussions among agents.

\subsection{Efficiency Study}
\begin{table}[htb!]
\centering
\caption{Model efficiency testing on GeoGlobe (Natural).}
\label{tab:exp_efficiency}
\renewcommand\tabcolsep{5pt}
\begin{filecontents*}{data/exp_efficiency.txt}
z;rt(avg);rt(mid);acc
0;473;473;75.2667
1;608;607;77.8349
3;766;754;81.5889
5;1,127;1,028;85.4603
10;1,897;1,123;86.2110
15;10,268;1,139;86.6456
20;22,176;1,289;87.0407
50;24,056;1,454;87.5543
\end{filecontents*}
\csvreader[
separator=semicolon,
  tabular=cccc,
  table head=\hline
 \bfseries{$Z$} & $RT_{\text{Avg}}$ (ms) & $RT_{\text{Med}}$ (ms) & $Acc$ (\%) \\
  \hline,
  late after last line= \\
  \hline % horizontal line at the end of the table
]{data/exp_efficiency.txt}{}{\csvlinetotablerow}
\end{table}

As illustrated in Table \ref{tab:exp_structure}, the proposed model achieves higher accuracy while consuming fewer tokens. To further demonstrate this efficiency, we introduce a more intuitive metric, \emph{Response Time} ($RT$), which reflects the overall performance of \name.

Our framework adapts to different application scenarios by utilizing the variable $Z$, as defined in stage 3 of the methodology section, to strike a balance between model accuracy and efficiency.
For example, when analyzing historical mobility patterns, a higher $Z$ value can be used since this task is less sensitive to response time but requires a higher accuracy. Conversely, for tasks such as recommending tourist attractions based on dynamic user posts, a lower $Z$ value is preferable to prioritize faster responses.

Table \ref{tab:exp_efficiency} reports the average and median of response times ($RT_{\text{Avg}}$, $RT_{\text{Med}}$), as well as the model accuracy ($Acc$). In general, as the maximum allowable discussion round $Z$ increases, the response time of \name\ also increases. However, thanks to implementing our dynamic learning model with an early stopping strategy, this growth remains nearly linear rather than exponential.

Then, the comparison of $RT_{\text{Avg}}$ and $RT_{\text{Med}}$ shows that only a small fraction of image queries require multiple discussion rounds. These outliers, with significantly longer response times, skew the average upwards. From the median response time, we observe that at least 50\% of the test samples receive a response within 2 seconds, an acceptable performance level for most applications.

\subsection{Parameter Analysis}

\begin{figure}[ht]
\centering
\subfigure[Calling the same LVLM API]{
\label{fig:exp_agent_num_a}
\resizebox{0.475\linewidth}{!}{
\begin{tikzpicture}
\begin{axis}[
	smooth,
	grid=major,
	xlabel=Number of Agents,
	ylabel=$Acc$ (\%),
	xmin=1,
	xmax=13,
        ymin=60,
        ymax=90,
	xtick = {1,3,5,7,9,11,13},
	legend columns=1,
	legend style={font=\normalsize,at={(0.25,0.02)},anchor=south, text opacity=1, fill opacity=0.5},
	font=\large
	]
    \addplot [mark=*,mycolor61, line width=1pt] table [x=x, y=Blender,, col sep=comma] {data/exp_agent_num.txt};
    \addplot [mark=square*,mycolor62, line width=1pt] table [x=x, y=PHP,, col sep=comma] {data/exp_agent_num.txt};
    \addplot [mark=triangle*,mycolor63, line width=1pt] table [x=x, y=Reflexion,, col sep=comma] {data/exp_agent_num.txt};
    \addplot [mark=pentagon*,mycolor64, line width=1pt] table [x=x, y=Debate,, col sep=comma] {data/exp_agent_num.txt};
    \addplot [mark=halfcircle*,mycolor65, line width=1pt] table [x=x, y=DyLAN,, col sep=comma] {data/exp_agent_num.txt};
    \addplot [mark=diamond*,mycolor66, line width=1pt] table [x=x, y=Ours,, col sep=comma] {data/exp_agent_num.txt};
    \legend{LLM-Blender,PHP,Reflexion,LLM Debate,DyLAN,\name}
\end{axis}
\end{tikzpicture}
}}
\subfigure[Different LVLM backbones]{
\label{fig:exp_agent_num_b}
\resizebox{0.475\linewidth}{!}{
\begin{tikzpicture}
\begin{axis}[
	smooth,
	grid=major,
	xlabel=Number of Agents,
	ylabel=$Acc$ (\%),
	xmin=1,
	xmax=13,
        ymin=20,
        ymax=90,
	xtick = {1,3,5,7,9,11,13},
	legend columns=1,
	legend style={font=\normalsize,at={(0.78,0.02)},anchor=south, text opacity=1, fill opacity=0.5},
	font=\large
	]
    \addplot [mark=*,mycolor61, line width=1pt] table [x=x, y=Blender,, col sep=comma] {data/exp_agent_num_b.txt};
    \addplot [mark=square*,mycolor62, line width=1pt] table [x=x, y=PHP,, col sep=comma] {data/exp_agent_num_b.txt};
    \addplot [mark=triangle*,mycolor63, line width=1pt] table [x=x, y=Reflexion,, col sep=comma] {data/exp_agent_num_b.txt};
    \addplot [mark=pentagon*,mycolor64, line width=1pt] table [x=x, y=Debate,, col sep=comma] {data/exp_agent_num_b.txt};
    \addplot [mark=halfcircle*,mycolor65, line width=1pt] table [x=x, y=DyLAN,, col sep=comma] {data/exp_agent_num_b.txt};
    \addplot [mark=diamond*,mycolor66, line width=1pt] table [x=x, y=Ours,, col sep=comma] {data/exp_agent_num_b.txt};
    \legend{LLM-Blender,PHP,Reflexion,LLM Debate,DyLAN,\name}
\end{axis}
\end{tikzpicture}
}}
\caption{Results of model performance in relation to the number of agents.}
\label{fig:exp_agent_num}
\end{figure}
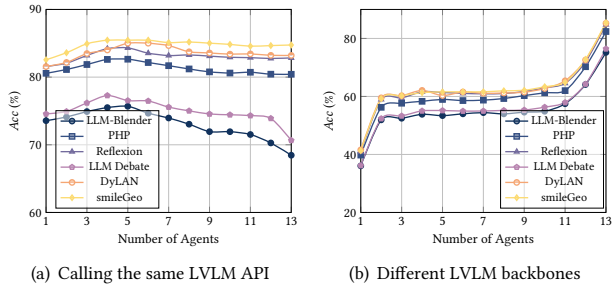

\noindent \textbf{Number of Agents}. We further demonstrate the relationships between the number of agents and the framework performance.
We conduct experiments in two ways: (i) by calling the same closed-source LVLM API (Here, we use Gemini-1.5-pro because it performs best without the help of the Internet) under different prompts (\eg You are good at recognizing natural attractions; You're a traveler around Europe) to simulate different agents, and (ii) by using different LVLM backbones to represent distinct agents.
The results are shown in Figure \ref{fig:exp_agent_num}.

As illustrated in Figure \ref{fig:exp_agent_num_a}, the framework achieves optimal accuracy with 4 or 5 agents.
Beyond this number, the framework's performance begins to deteriorate.
This shows that using models with the same knowledge and reasoning capabilities as different agents has limited improvement in the accuracy of the framework.
Despite this decline, the performance of frameworks other than LLM-Blender and LLM Debate remains superior to that of a single agent.
LLM-Blender and LLM Debate, however, have a significant decrease in model accuracy when the number of agents exceeds 11.
This is mainly because both of them involve all LVLMs in every discussion, which suffers from excessive repetitive and redundant discussions.
Figure \ref{fig:exp_agent_num_b} reveals that the accuracy of the framework improves with the incorporation of more LVLM backbones, indicating that the diversity of LVLMs can enhance the quality of discussions.

% \begin{wrapfigure}{r}{0.43\textwidth} % 'r' 代表右对齐，'0.5\textwidth' 表示图形宽度为文本宽度的一半
\begin{figure}[!ht]
\centering
\resizebox{0.75\linewidth}{!}{
\begin{tikzpicture}
    \begin{axis}[
        view={0}{90},   % not needed for `matrix plot*' variant
        xlabel=$R$,
        xtick = {0,1,2,3,4,5,6,7},
        xticklabels = {1,2,3,4,5,6,7,8},
        ylabel=$K$,
        ytick = {0,1,2,3,4,5,6,7},
        yticklabels = {1,2,3,4,5,6,7,8},
        colormap={mycolormap}{
            % color=(mycolor01),
            % color=(mycolor02),
            color=(mycolor03),
            color=(mycolor04),
            color=(mycolor05),
            color=(mycolor06),
            color=(mycolor07),
            color=(mycolor08),
            color=(mycolor09),
            color=(mycolor10),
            color=(mycolor11),
            color=(mycolor12),
            color=(mycolor13),
            % color=(mycolor14),
            % color=(mycolor15),
            % color=(mycolor16)
        },
        colorbar,
        colorbar style={
            title=Acc (\%),
            yticklabel style={
                /pgf/number format/.cd,
                precision=0,
                fixed zerofill,
            },
        },
        % title=data from infrared measurements,
        enlargelimits=false,
        axis on top,
        point meta min=82,
        point meta max=86,
    ]
        \addplot [matrix plot*,point meta=explicit] file  {data/heat.dat};
    \end{axis}
\end{tikzpicture}
}
\caption{Results under different $K$ and $R$.}
\label{fig:heat}
\end{figure}
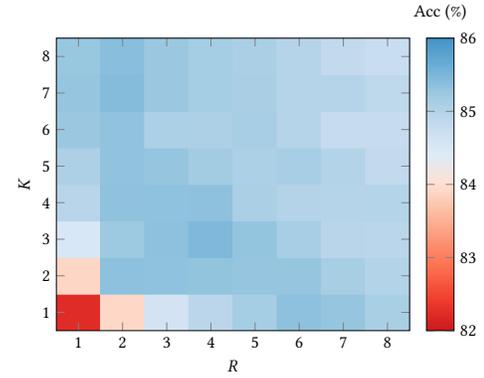
% \end{wrapfigure}

\noindent \textbf{Hyperparameter $K$ \& $R$}.
There are two hyperparameters, $K$ and $R$, that need to be pre-defined in the proposed framework:
$K$ is the number of agents (answer agents) that respond in each round of discussion, and $R$ is the number of agents (review agents) used to review answers from answer agents.
Therefore, we conduct experiments under different combinations of $K \in [1,8]$ and $R \in [1,8]$, as shown in Figure \ref{fig:heat}. 
The results indicate that optimal performance can be achieved with relatively small values of 
$K$ or $R$.
However, the computational cost, measured in tokens, increases exponentially with higher values of $K$ and $R$.
To ensure the accuracy of the \name\ while reducing the calculation cost as much as possible, we set both $K$ and $R$ equal to 2 in this paper.

\subsection{Case Study}

To further illustrate the superiority of our proposed framework, we provide detailed examples. Additional information about the case study is presented in Appendix \ref{app:case}.

\section{Conclusion}
\label{sec:cc}

This work introduces a novel LVLM agent framework, \name, specifically designed for geo-localization tasks. Inspired by the review mechanism, it integrates various LVLMs to discuss anonymously and geo-localize images worldwide. Additionally, we have developed a dynamic learning strategy for agent collaboration social networks, electing appropriate agents to geo-localize each image with different characteristics. This enhancement reduces the computational burden associated with collaborative discussions among LVLM agents.
Moreover, we have constructed a geo-localization dataset called GeoGlobe to evaluate the proposed framework better.
Overall, \name\ demonstrates significant improvements in geo-localization tasks, achieving superior performance with lower computational costs compared to state-of-the-art baselines.

Looking ahead, we aim to expand the capabilities of \name\ to incorporate more powerful external tools beyond just web searching. Additionally, we plan to explore extending its application to more complex scenarios, such as high-precision global positioning,
% and navigation for robots, 
laying the cornerstone for exploring LVLM agent collaboration to handle different complex open-world tasks efficiently. More discussions are detailed in Appendix \ref{app:discussion}.

% \begin{acks}
% This research was partially supported by APRC - CityU New Research Initiatives (No.9610565, Start-up Grant for New Faculty of City University of Hong Kong), CityU - HKIDS Early Career Research Grant (No.9360163), SIRG - CityU Strategic Interdisciplinary Research Grant (No.7020046, No.7020074), Tencent (CCF-Tencent Open Fund), Huawei (Huawei Innovation Research Program) and Ant Group (CCF-Ant Research Fund, Ant Group Research Fund)
% , InnoHK initiative, the Government of the HKSAR, and the Laboratory for AI-Powered Financial Technologies.
% \end{acks}

%\bibliographystyle{ACM-Reference-Format}
%\bibliography{90Reference}


\appendix
\newpage

\section{Notations}

We summarize all notations in this paper and list them in Table \ref{tab:notation}.

\begin{table}[htbp!]
\centering
\caption{Notations in this paper.}
\label{tab:notation}
\renewcommand\tabcolsep{2.5pt}
\begin{filecontents*}{data/notation.txt}
Notation; Description
$\boldsymbol{X}$; The image to be recognized.
$\boldsymbol{Y}$ ($\boldsymbol{\hat{Y}}$); The predicted (ground truth of) location.
$\mathcal{G}$ ($\mathcal{\hat{G}}$); The predicted (ground truth of) LVLM social network.
$\boldsymbol{A}$ ($\boldsymbol{\hat{A}}$); The predicted (ground truth of) adjacency matrix.
$\boldsymbol{Lst}$ ($\boldsymbol{\hat{Lst}}$); The predicted (ground truth of) probability list.
$\mathcal{V}$; The set of LLM agents.
$\mathcal{E}$; The set of connections between LLM agents.
$N$; The number of agents.
$K$; The number of answer agent(s).
$R$; The number of review agent(s).
$L$; The number of agent discussion rounds.
$Z$; The maximum number of discussion rounds.
$\Theta$; The learnable parameters in agent selection model.
\end{filecontents*}
\csvreader[
separator=semicolon,
  tabular=cc,
  table head=\hline
 \multicolumn{1}{c}{\bfseries{Notation}} & \multicolumn{1}{c}{\bfseries{Description}} \\
  \hline,
  late after last line= \\
  \hline % horizontal line at the end of the table
]{data/notation.txt}{}{\csvlinetotablerow}
\end{table}

\section{Dataset Details}
\label{app:dataset}

\begin{figure}[htbp!]
\centering
\includegraphics[width=\linewidth]{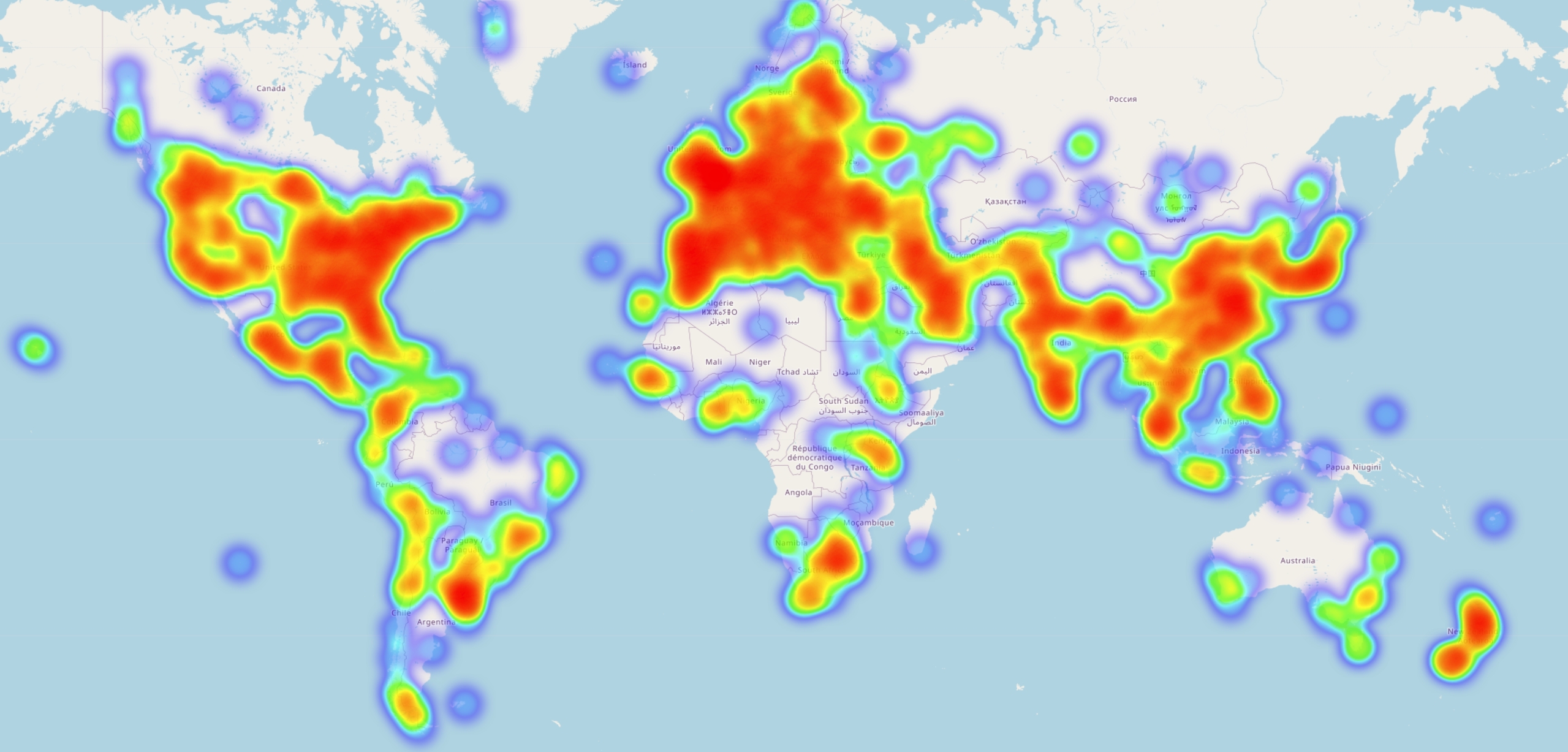} % 插入图片，调整为minipage宽度
\caption{The data distribution around the world.}
\label{fig:geodis}
\end{figure}

As a supplement to the two open source datasets, we also constructed a new dataset, GeoGlobe. The images in this dataset are copyright-free images obtained from the Internet via a crawler.
We divide the images into two main categories: man-made landmarks and natural attractions.
Then, we filter out the data samples that could clearly identify the locations of the landmarks or attractions in the images.
As a result, we filter out nearly three hundred thousand data samples, and please refer to Table \ref{tab:datasets} and Figure \ref{fig:geodis} for details.
Due to the fact that a large number of natural attractions in different geographical regions with high similarity are cleaned, the magnitude of the data related to natural attractions in this dataset is smaller than that of man-made attractions.

% \noindent % 确保没有缩进干扰布局

% \hfill % 添加必要的间隔

\begin{table}[t]
\centering
% \vspace{-52pt}
\caption{Statistics of the dataset GeoGlobe.}
\label{tab:datasets}
\renewcommand\tabcolsep{2.5pt}
\begin{tabular}{ccccc} % 开始一个表格
\hline
& \bfseries{Images} & \bfseries{Cities} & \bfseries{Countries} & \bfseries{Attractions} \\ \hline
Man-made & 253,118 & 2,313 & 143 & 10,492 \\
Natural & 40,087 & 1,044 & 97 & 1,849 \\ \hline
\end{tabular}
\end{table}

For an open-world geo-localization task, the relationship between the training and test samples in the experiment could greatly affect the results.
We label the training samples as $\mathcal{Z}_{\text{train}}$, and the test sample set as $\mathcal{Z}_{\text{test}}$, and use two metrics, \emph{coverage} as well as \emph{consistency}, to portray this relationship:

\begin{equation}
\label{eq:sample_relation}
\begin{aligned}
coverage &= \frac{\mathcal{Z}_{\text{train}} \cap \mathcal{Z}_{\text{test}}}{\mathcal{Z}_{\text{train}}} \times 100\% \\
consistency &= \frac{\mathcal{Z}_{\text{train}} \cap \mathcal{Z}_{\text{test}}}{\mathcal{Z}_{\text{test}}} \times 100\%
\end{aligned}
\end{equation}

As for the samples in this paper, $coverage \approx 4.6564\%$, and $consis-$ $tency$ $\approx 33.2957\%$.

\section{Implementation Details}
\label{app:implementation}

\begin{algorithm}[htbp!]
\caption{The \name\ framework}
\label{alg:train}
\begin{algorithmic}[1]

\REQUIRE A set of pre-trained LLMs $\mathcal{V} = \left\{ v_1, v_2, \cdots \right\}$, the input image $\boldsymbol{X}$, and the ground truth $\boldsymbol{\hat{Y}}$ (if has);
\ENSURE The geospatial location $\boldsymbol{Y}$.

\emph{Initialization Stage:}

\STATE Initialize (Load) the parameter of the agent selection model: $\Theta$

\STATE Calculate: $\boldsymbol{A} \leftarrow f(X, \mathcal{V} | \Theta)$

\STATE Initialize the agent collaboration social network: $\mathcal{G}$

\STATE Calculate: $\boldsymbol{Lst} \leftarrow f(X, \mathcal{G} | \Theta)$

\emph{Stage 1:}

\STATE Elect $K$ answer agents: $\mathcal{V}^1 = \{v_a^1, v_b^1, \cdots \} \leftarrow \boldsymbol{Lst}$

\FOR{each answer agent $v^1$}

\STATE Obtain the location: $\boldsymbol{Y}_{v^1}^1 \leftarrow \mathrm{Ask}_{v^1}(\boldsymbol{X})$

\STATE Get the confidence percentage: $C_{v^1}^1 \leftarrow \mathrm{Ask}_{v^1}(\boldsymbol{X}, \boldsymbol{Y}_{v^1}^1)$

\STATE Store the further explanation: $T_{v^1}^1 \leftarrow \mathrm{Ask}_{v^1}(\boldsymbol{X}, \boldsymbol{Y}_{v^1}^1)$

\ENDFOR

\emph{Stage 2:}

\FOR{each selected answer agent $v^1$}

\STATE Select $R$ review agents: \\
\quad $\mathcal{V}^2 = \{v_a^2, v_b^2, \cdots \} \leftarrow \mathrm{RandomWalk}_{v^1}(\mathcal{G})$

\FOR{each review agent $v^2$}

\STATE Obtain the comment $T_{v^2}^2 \leftarrow \mathrm{Review}_{v^2}(\boldsymbol{X}, \boldsymbol{Y}_{v^1}^1, C_{v^1}^1)$

\STATE Get the confidence percentage: $C_{v^2}^2 \leftarrow \mathrm{Ask}_{v^2}(\boldsymbol{X}, T_{v^2}^2)$

\ENDFOR

\ENDFOR

\emph{Stage 3:}

\FOR{each selected answer agent $v^1$}

\STATE Summary the final answer: \\
\quad $\boldsymbol{Y}^3_{v^1} \leftarrow \mathrm{Summary}_{v^1}(\boldsymbol{Y}_{v^1}^1, C_{v^1}^1, T^2_{v^2_1}, C_{v^2_1}^2, T^2_{v^2_2}, , C_{v^2_2}^2, \cdots)$

\STATE Get the final confidence percentage: \\
\quad $C^3_{v^1} \leftarrow \mathrm{Ask}_{v^1}(\boldsymbol{Y}_{v^1}^1, C_{v^1}^1, T^2_{v^2_1}, C_{v^2_1}^2, T^2_{v^2_2}, , C_{v^2_2}^2, \cdots)$

\ENDFOR

\STATE Generate the final answer: \\
\quad $\boldsymbol{Y} \leftarrow \mathrm{Discussion}_Z(\boldsymbol{Y}^3_{v^1_1}, C^3_{v^1_2}, \boldsymbol{Y}^3_{v^1_2}, C^3_{v^1_2}, \cdots)$

\emph{The dynamic learning strategy module:}

\STATE Initialize $\boldsymbol{Lst}^{(0)}, \mathcal{G}^{(0)}$

\FOR{round $l$ in total $L$ rounds}

\FOR{each selected answer agent $v^1$}

\STATE Obtain coordinates: \\
\quad $Coors \leftarrow \mathrm{GeoEmb}(\boldsymbol{Y}^3_{v^1})$, \\
\quad $Coors_{\text{Truth}} \leftarrow \mathrm{GeoEmb}(\boldsymbol{Y}_{\text{Truth}})$

\IF{$\mathrm{Dis}(Coors, Coors_{\text{Truth}}) \leq th$}

\STATE $\boldsymbol{A}^{(l)} \leftarrow \mathrm{Enhance}(e | e \text{ contains } v^1, e \in \mathcal{E})$

\STATE Update $\boldsymbol{Lst}^{(l)}[v^1] = 1$

\ELSE

\STATE $\boldsymbol{A}^{(l)} \leftarrow \mathrm{Weaken}(e | e \text{ contains } v^1, e \in \mathcal{E})$

\STATE Update $\boldsymbol{Lst}^{(l)}[v^1] = 0$

\ENDIF

\ENDFOR

\ENDFOR

\STATE $\boldsymbol{\hat{A}} \approx \boldsymbol{A}^{(L)}$, $\boldsymbol{\hat{Lst}} \approx \boldsymbol{Lst}^{(L)}$

\STATE Update: $\Theta \leftarrow Loss(\boldsymbol{\hat{Y}}, \boldsymbol{Y}, \boldsymbol{\hat{A}}, \boldsymbol{A}, \boldsymbol{\hat{Lst}}, \boldsymbol{Lst})$

\end{algorithmic}

\end{algorithm}

In all experiments, we employ a variety of LVLMs, encompassing both open-source and closed-source models, to be agents in the proposed framework.
Unless specified otherwise, zero-shot prompting is applied.
Each open-source LVLM is deployed on a dedicated A800 (80G) GPU server with 200GB memory.
As for each closed-source LVLM, we cost amounting to billions of tokens by calling APIs as specified by the official website.
To avoid the context length issue that occurs in some LVLMs, we truncate the context before submitting it to the agent for questions based on the maximum number of tokens that each agent supports.
Besides, noting that images are token consuming, we only keep the freshest response for discussions among different agents.

\begin{figure*}
    \centering
    \includegraphics[width=0.94\linewidth]{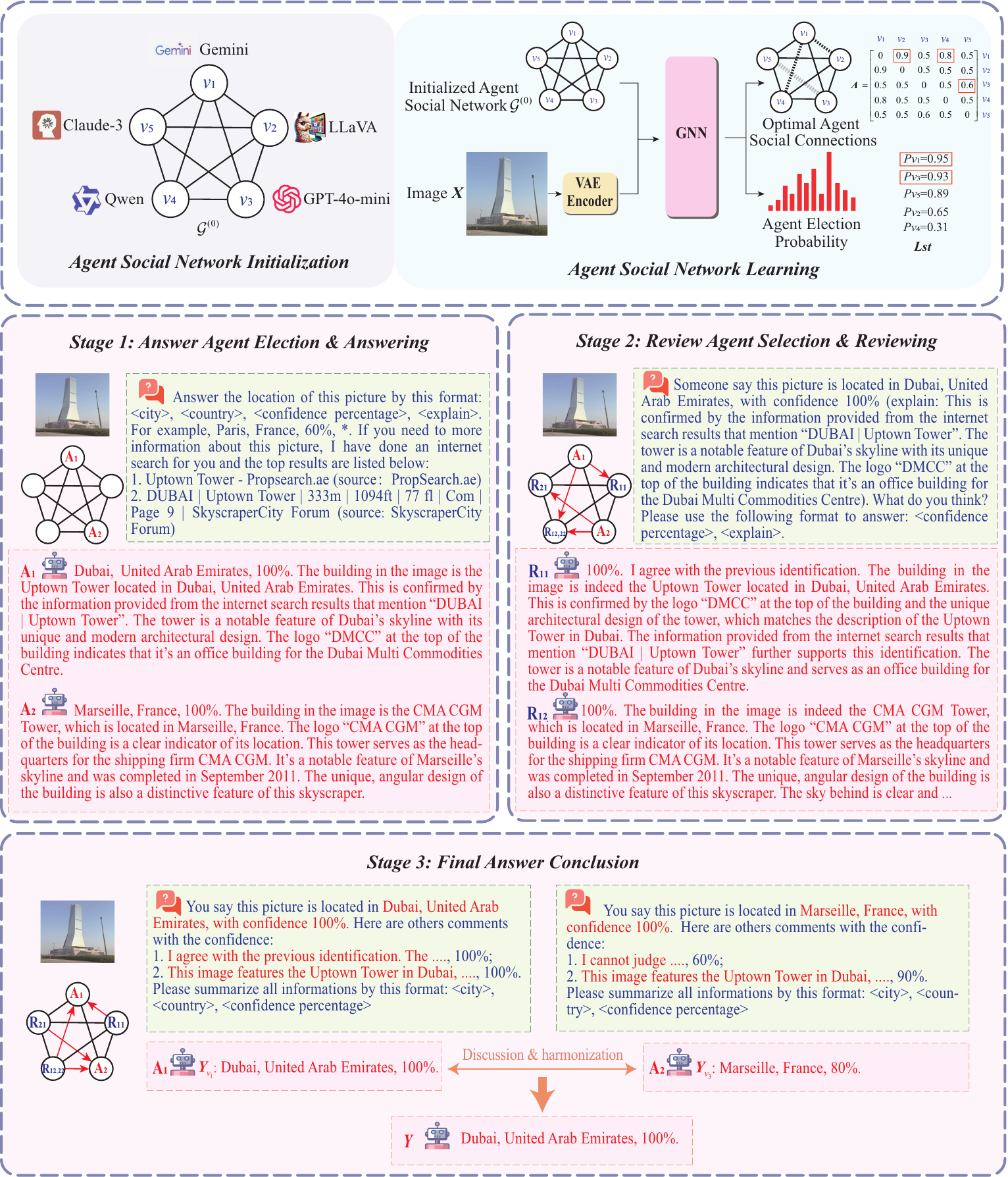}
    \caption{
    A case study on the geo-localization process via a given image.
    }
    \label{fig:case}
\end{figure*}

The detailed algorithm of \name\ is illustrated in Algorithm \ref{alg:train}.
In the initialization stage, we initialize or load the parameters of the agent social network learning model, as delineated in line 1.
Next, we treat each LVLM agent as a node, establishing the LVLM agent collaboration social network and computing the adjacency relationships among LVLM agents as well as the probability that each agent is suited for responding to image $\boldsymbol{X}$, as shown in line 2.
Then, line 3 initializes the agent collaboration social network and line 4 computes the agent election probability.
In Stage 1, line 5 involves electing appropriate answer agents based on the calculated probabilities.
Subsequently, lines 6-10 detail the process through which each chosen answer agent formulates their response.
Stage 2 begins by employing the random walk algorithm to assign review agents to each answer agent, as depicted in lines 11-12.
Lines 13-16 then describe how these review agents generate feedback based on the answers provided.
In Stage 3, each answer agent consolidates feedback from their assigned review agents to finalize their response, as illustrated in lines 18-21.
Line 22 concludes the final answer with up to $Z$ rounds (we set $Z=10$ in experiments) of intra-discussion among all answer agents only.
The dynamic learning strategy module involves $L$-round (we set $L=20$ in experiments) comparing the generated answers against the ground truth and updating the connections between the answer and review agents accordingly, as shown in lines 23-36.
In line 37, the process concludes with the updating of the learning parameters of the dynamic agent social network learning model.

Here, for the agent social network learning model, we first deflate each image to be recognized to 512x512 pixels and then use the pre-trained VAE model\footnote{\href{https://huggingface.co/stabilityai/sd-vae-ft-mse}{https://huggingface.co/stabilityai/sd-vae-ft-mse}} to compress the image again (compression ratio 1:8) and extract its representations.
We define the embedding dimension of the nodes to be 1024 and the hidden layer dimension of the network layer to be 1024.
we use Adam as an optimizer for gradient descent with a learning rate of $1e^{-5}$.
For each stage of the LVLM agent discussion, we use a uniform template to ask questions to different LVLM agents and ask them to respond in the specified format.
In addition, the performance of our proposed framework is the average of the last 100 epochs in a total training of 2500 epochs.

\section{Case Study}
\label{app:case}

\noindent \textbf{Case 1:} In Figure \ref{fig:case}, we illustrate the application of \name\ in a visual geo-localization task.
For this demonstration, we randomly select an image from the test dataset and employ five distinct LVLMs: LLaVA, GPT-4o-mini, Claude-3-opus, Gemini-1.5-pro, and Qwen2.
The agent selection model selects two answer agents, as depicted in the top part of the figure. Subsequently, stages 1 through 3 detail the process of generating the accurate geo-location.
Initially, only one answer agent provided the correct response.
However, after several rounds of discussion, the agent that initially responded incorrectly revised the confidence level of its answer.
During the final internal discussion, this agent aligned its response with the correct answer.
This outcome validates the efficacy of our proposed framework, demonstrating its ability to integrate the knowledge and reasoning capabilities of different agents to enhance the overall performance of the proposed LVLM agent framework.

\begin{figure*}[htbp!]
\centering
\subfigure[An example of the input image]{
\includegraphics[width=0.35\linewidth]{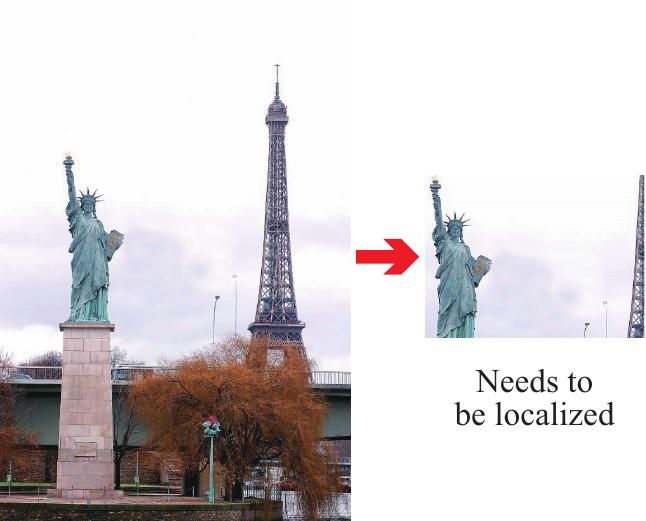}
\label{fig:cs2a}
}
\subfigure[Various thoughts about the image]{
\makebox[0.45\linewidth][c]{
\includegraphics[width=0.45\linewidth]{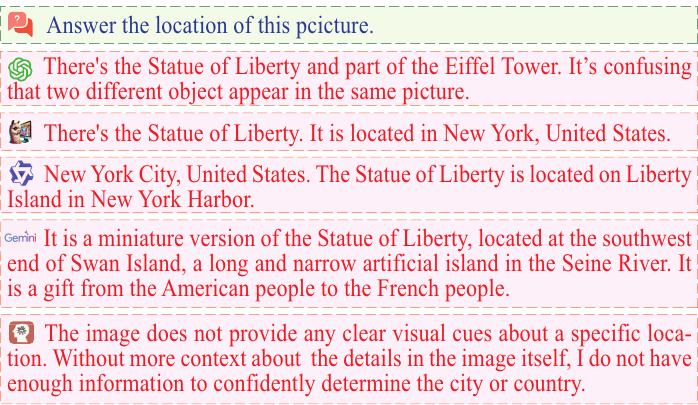}}
\label{fig:cs2b}
}

\subfigure[Actual locations of two landmarks]{
\includegraphics[width=0.33\linewidth]{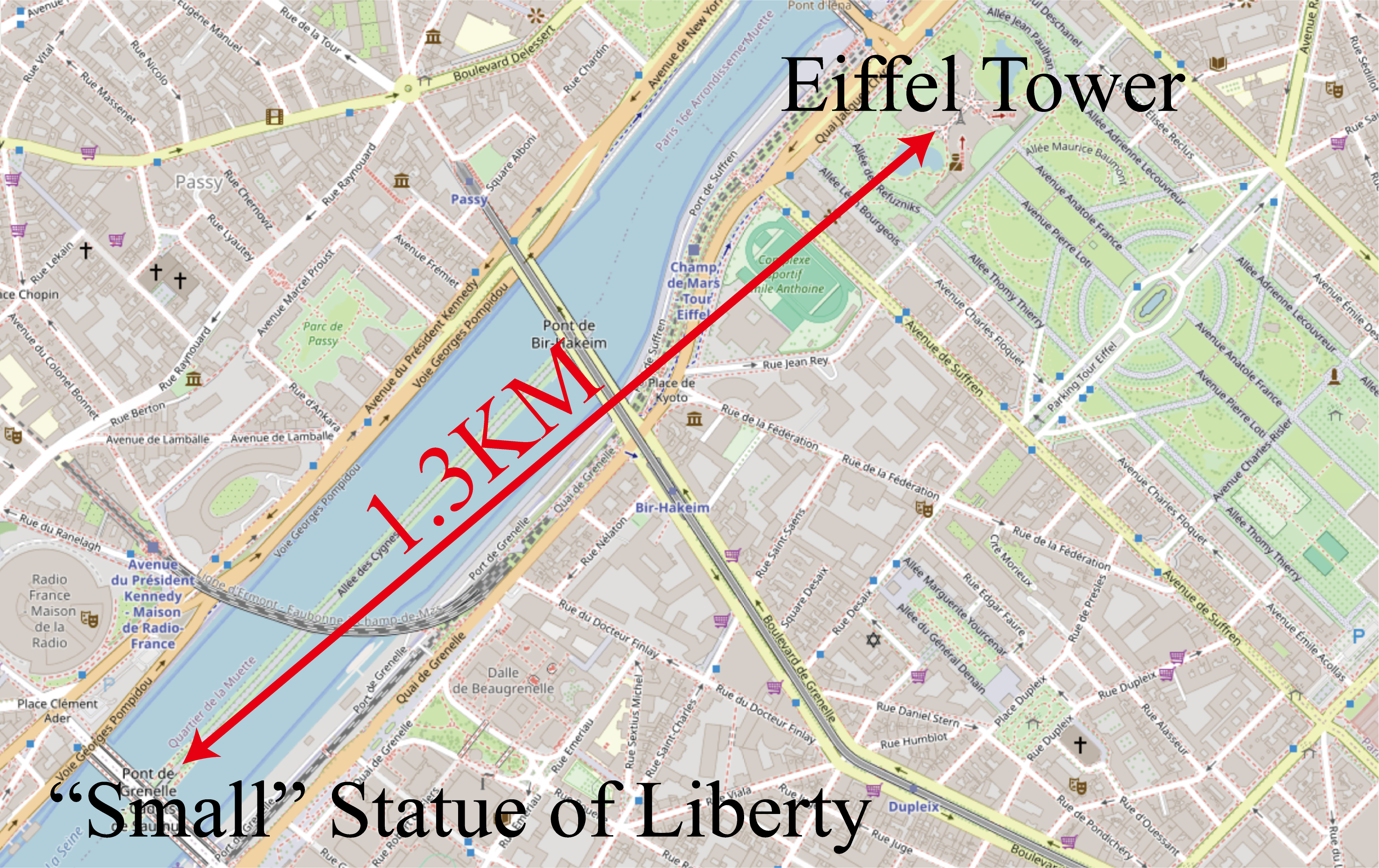}
\label{fig:cs2c}
}
\subfigure[The final answer of \name]{
\includegraphics[width=0.47\linewidth]{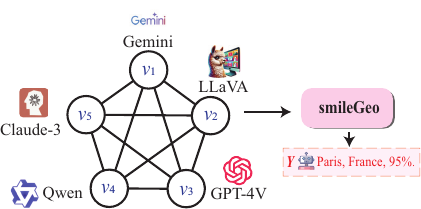}
\label{fig:cs2d}
}

\caption{A case study illustrating the reasoning capabilities of \name.}
\label{fig:3}
\end{figure*}

\noindent \textbf{Case 2:} This case study illustrates the need to pinpoint the geographical location of a complete image based on only a portion of it, as demonstrated in \ref{fig:cs2a}.
As illustrated in Figure \ref{fig:cs2b}, all agents recognized the Statue of Liberty in Figure \ref{fig:cs2a}, and some identified the presence of part of the Eiffel Tower at the edge of the picture.
For instance, GPT-4o mini concluded that the buildings in these two locations appeared in the same image.
However, as is known through the knowledge of other agents (Gemini), a scaled-down version of the Statue of Liberty has been erected on Swan Island, an artificial island in the Seine River in France.
By marking both the Eiffel Tower and the island on the Open Street Map (OSM) manually, as shown in Figure \ref{fig:cs2c}, it is evident that they are merely 1.3 kilometers apart in a straight line.
By utilizing the proposed framework, agents discuss and summarize the location depicted in the picture to be Paris, France, as shown in Figure \ref{fig:cs2d}.
Thus, without human intervention, this framework demonstrates the effectiveness of doing geo-localization tasks.

\section{Discussion}
\label{app:discussion}

\begin{figure*}[htbp!]
\centering
\subfigure[List of original Internet search results]{
\includegraphics[width=0.45\linewidth]{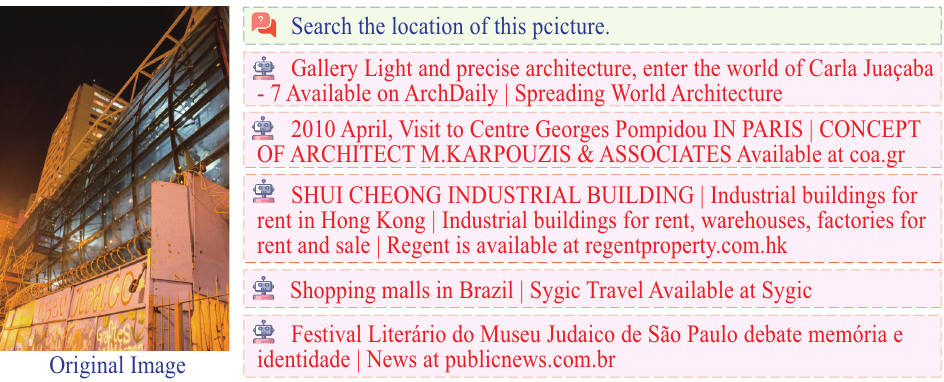}
\label{fig:discussion_a}
}
\subfigure[List of results after manual screening]{
\makebox[0.45\linewidth][c]{
\includegraphics[width=0.45\linewidth]{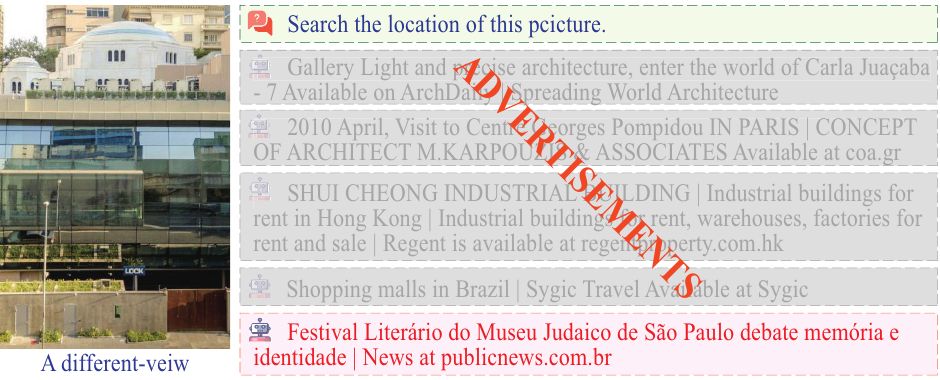}}
\label{fig:discussion_b}
}
\caption{A case study demonstrating retrieval results can interfere with the performance.}
\label{fig:discussion}
\end{figure*}

The proposed framework demonstrates strong performance in geo-localization, with experimental results showing that SmileGeo significantly outperforms other models and systems that rely purely on image similarity retrieval. However, there remain opportunities for future improvement. One notable area for enhancement is the susceptibility to interference when the number of agents is small. We illustrate this issue with an example, as depicted in Figure \ref{fig:discussion}.

As shown in Figure \ref{fig:discussion_a}, the data retrieved from the internet frequently includes a significant amount of irrelevant advertising information.
This extraneous content hampers the model's accuracy in image localization.
Figure \ref{fig:discussion_b} further reveals that, after human workers manually filtered out the advertising content, 80\% of the retrieval results were unrelated to the queried image.
This underscores the necessity of enhancing the model's resilience to such interference, which will be a major focus of our future research.

Another critical challenge is the computational efficiency, which is a common limitation of LVLMs. Currently, a range of methods exists, such as model distillation and compression, which can enhance computational efficiency. For instance, in our current implementation, we used high-performance servers located in U.S. data centres to support the system and deployed smaller models, like GPT-4o-mini, to reduce response times. According to our experiments, GPT-4o-mini demonstrated substantially improved response times compared to previous models, such as GPT-4V. We anticipate that ongoing advancements in large-model technology will continue to address these efficiency concerns.
In the future, we plan to incorporate distillation-based models as agents within our framework, aiming to enhance both computational performance and robustness.
% By leveraging model distillation, we can maintain the essential capabilities of large models while significantly reducing their size, resulting in a more lightweight and efficient system.
It is expected to improve system response time and resilience against interference further by enabling faster decision-making and better adaptability to noisy or irrelevant information.

\end{document}